\pdfoutput=1
\documentclass[11pt]{article}
\usepackage{booktabs}
\usepackage{acl}

\usepackage{times}
\usepackage{latexsym}
\usepackage{amsmath}
\usepackage{array}
\usepackage{tabularx}
\usepackage{multirow}
\usepackage{graphicx}
\usepackage{fontawesome}
\usepackage{academicons}
\usepackage{hyperref}
\hypersetup{
  colorlinks=true,
  citecolor=blue,      
  linkcolor=magenta,     
  urlcolor=blue        
}

\renewcommand{\arraystretch}{1.3}

\usepackage[T1]{fontenc}
\usepackage[utf8]{inputenc}
\usepackage{microtype}
\usepackage{inconsolata}

\usepackage{graphicx}
\usepackage[capitalise]{cleveref}
\usepackage[most]{tcolorbox}  

\usepackage{listings}        
\usepackage{xcolor}          
\usepackage{fancyvrb}
\definecolor{darkorange}{HTML}{0845ff}
\definecolor{softgray}{RGB}{245,246,250}
\definecolor{softblue}{RGB}{240,244,255}
\definecolor{darkblue}{RGB}{50,50,80}
\tcbset{
    mycustombox/.style={
        enhanced,  
        breakable, 
        colback=softgray, 
        colframe=gray!20, 
        coltitle=black, 
        fonttitle={\fontsize{11pt}{13pt}\bfseries\sffamily\small}, 
        rounded corners, 
        title=#1, 
        boxsep=4pt,          
        arc=2pt,
        before skip=1em,
        after skip=1em,
        fontupper=\fontsize{9pt}{12pt}\selectfont 
    },
    mycustombox_full/.style={
        enhanced,            
        breakable,           
        float*=htb, 
        floatplacement=htb, 
        width=\textwidth, 
        colback=gray!8, 
        colframe=darkgray, 
        coltitle=white, 
        fonttitle={\fontsize{11pt}{13pt}\bfseries\selectfont}, 
        rounded corners, 
        title=#1, 
        boxsep=4pt,          
        arc=2pt,
        before skip=1em,
        after skip=1em,
        fontupper=\fontsize{9pt}{12pt}\selectfont 
    },
    professionalbox/.style={
      enhanced,
      breakable,
    
      colback=softgray,
      colframe=gray!20,
    
      coltitle=black,
      fonttitle=\bfseries\sffamily\small,
    
      attach boxed title to top center={
        yshift=-2mm
      },
    
      boxed title style={
        colback=gray!20,
        colframe=gray,
        boxrule=0pt,
        bottom=0pt,
    
        width=\linewidth,
        halign=center,
      },
    
      title=#1,
    
      fontupper=\small,
      boxsep=5pt,
      left=5pt,
      right=5pt,
      top=10pt,
      arc=3pt,
      before skip=1.5em,
      after skip=1.5em
    },
    professionalbox_full/.style={
        professionalbox={#1},  
        float*=t,              
        width=\textwidth,      
    }
}
\newtcolorbox{professionalbox}[1]{%
  professionalbox/.style,
  title={#1}
}

\newcommand{\tfact}{t_{\mathrm{fact}}}
\newcommand{\tfactset}{S_{\mathrm{fact}}}
\newcommand{\cfactset}{S_{\mathrm{cofa}}}
\newcommand{\cfact}{t_{\mathrm{cofa}}}

\title{When Seeing Overrides Knowing: Disentangling Knowledge Conflicts \\ in Vision-Language Models}

\author{
Francesco Ortu$^{1,2}$ \quad Zhijing Jin$^{3,4,5}$\quad Diego Doimo$^2$\textsuperscript{\dag} \quad Alberto Cazzaniga$^{2}$\textsuperscript{\dag}  \\ [10px]
        $^1$University of Trieste \   $^2$AREA Science Park  \\ 
     $^3$MPI \ $^4$University of Toronto \ $^5$Vector Institute     
}

\begin{document}
\maketitle
\begingroup
\renewcommand\thefootnote{}
    \footnotetext{\hangindent=1.8em Correspondence: \texttt{\{francesco.ortu, diego.doimo, alberto.cazzaniga\}@areasciencepark.it}}
    \footnotetext{\hangindent=8.2em Code and Dataset: \faGithub~\href{https://github.com/francescortu/Seeing-Knowing}{francescortu/Seeing-Knowing}}
    \footnotetext{$^\dagger$ Equal supervision.}
\endgroup

\begin{abstract}
Vision-language models (VLMs) increasingly combine visual and textual information to perform complex tasks. However, conflicts between their internal knowledge and external visual input can lead to hallucinations and unreliable predictions. 
In this work, we investigate the mechanisms that VLMs use to resolve cross-modal conflicts by introducing \texttt{\textsc{WHOOPS-AHA!}}, a dataset of multimodal counterfactual queries that deliberately contradict internal commonsense knowledge.
Through logit inspection, we identify a small set of attention heads that mediate this conflict. 
By intervening in these heads, we can steer the model towards its internal parametric knowledge or the visual information. 
Our results show that attention patterns on these heads effectively locate image regions that influence visual overrides, providing a more precise attribution compared to gradient-based methods.
\end{abstract}

\section{Introduction}
Vision–language models (VLMs) \cite{flamingo, blip, llava_nips, team2024chameleon, molmo2024} have shown remarkable versatility in various multimodal tasks, from image understanding to image generation.
They draw on their ability to combine two key sources of information:
a rich world knowledge acquired during pretraining, and contextual cues provided in the input prompts. 
However, these two sources can sometimes contradict each other, for example when the pretraining knowledge becomes outdated \cite{lazaridou2021mind, luu-etal-2022-time} or when prompts include intentionally misleading visual information \cite{liu2024promptinjectionattackllmintegrated}. 
Such conflicts often lead to hallucinations in model responses \cite{DBLP:journals/corr/abs-2311-03287, liu2024mitigating, guan2024hallusionbenchadvanceddiagnosticsuite}, 
yet the internal mechanisms responsible for these errors remain poorly understood \cite{xu-etal-2024-knowledge-conflicts}.

\begin{figure}[t]
        \centering
        \includegraphics[width=0.99\linewidth]{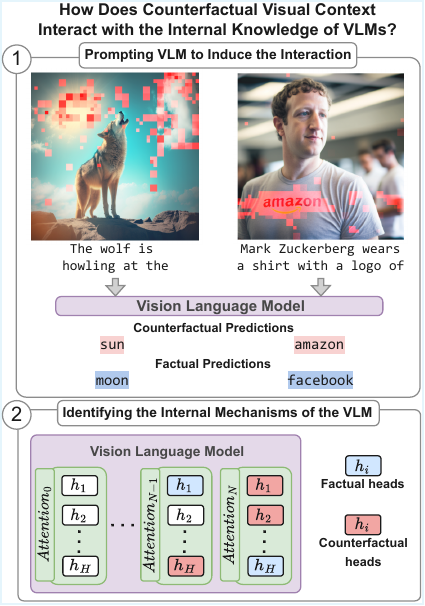}
        \vspace{-0.6cm}
        \caption{\textbf{Overview of our approach.} (\textbf{Top}) We construct prompts that induce a conflict between a vision-language model’s internal factual knowledge and counterfactual visual context. (\textbf{Bottom}) We then analyze which components in the model mediate this tension, identifying attention heads and visual patches that favor factual or visually grounded predictions.}
        \label{fig:cartoon}
    \vspace{-0.5cm}
    \end{figure}

In this work, we analyze how VLMs resolve conflicts between visual input and internal knowledge by framing the problem through counterfactual image-text pairs.
We prompt the VLMs with images depicting unusual or absurd scenes taken from the \texttt{WHOOPS!} dataset \cite{Bitton2023whoops}, followed by a sentence encouraging a typical knowledge-based continuation. 
As shown in \cref{fig:cartoon}, each input prompt is associated with a counterfactual pair of completions. For instance, the model may be shown an image of a wolf howling at the sun, a scene that contradicts commonsense knowledge, and asked to complete the prompt accordingly (see top-left panel). 
We construct the dataset such that VLMs generate commonsense responses when prompted with text alone, but override them in the presence of an image to align with the visual context, even when it contradicts their internal knowledge.
Building on the approach of \citet{ortu2024competition}, we identify which internal components of the model contribute the most to factual (inner knowledge) versus counterfactual (image context) predictions. 
We find that a small subset of attention heads mediates this competition, and targeted interventions on these heads can reliably alter the model’s outputs. 
We also show that these heads are more effective than gradient-based methods in identifying the most important parts of an image to resolve multimodal conflicts in VLMs.
In summary, our contributions are as follows:
\begin{enumerate}
    \item We construct \texttt{\textsc{WHOOPS-AHA!}}, a dataset that combines images containing counterfactual scene elements and commonsense textual queries, designed to analyze conflicts between visual context and internal knowledge (\cref{sec:whoops});
    \item We identify the attention heads that promote factual and counterfactual responses, ranking their importance with logit attribution (\cref{sec:attention_heads});
    \item By reweighting these heads, we show that we can control the tendency of the model to rely on the visual evidence or its internal knowledge and vice versa (\cref{sec:intervention});
    \item We demonstrate that direct attention attribution from conflict-resolution heads provides more accurate identification of counterfactual image regions than traditional gradient-based attribution methods
    (\cref{sec:attention_attribution}).
\end{enumerate}

\section{Related Work}
\paragraph{Knowledge conflicts in VLMs.}
VLMs frequently encounter situations where visual input contradicts their internal parametric knowledge, yet the mechanisms governing conflict resolution remain poorly understood \citep{xu-etal-2024-knowledge-conflicts}. Early work on multimodal conflicts focused primarily on behavioral evaluation through benchmark construction. \citet{han-etal-2024-instinctive} introduced datasets that probe contextual knowledge conflicts with deceptive visual elements, while \citet{liu2024insightsightexploringvisionknowledge} developed ConflictVis to evaluate conflicts between visual input and parametric knowledge. \citet{Le2023cococounterfact}  created \texttt{\textsc{COCO-Counterfactuals}} using minimally edited counterfactual image pairs to study model behavior under visual contradictions. More recently, \citet{zhu2024unraveling} formally characterize cross-modality parametric knowledge conflict in VLMs, showing that 
conflict rates remain persistently high regardless of model size. However, these studies limit their analysis to evaluating model outputs and prompt structures without investigating the internal mechanisms by which models resolve conflicts.

\paragraph{Mechanistic interpretability in VLMs.}
Mechanistic interpretability, which seeks to reverse engineer deep neural networks, has made significant strides in text-only models \citep[inter-alia]{elhage2021mathematical, geva2023dissecting, geva2021keyvalue, Hanna2023greater-then}. Recently, attention has shifted to VLMs. Early work adapted tools from the language setting to multimodal architectures: \citet{schwettmann2023multimodal} identified MLP neurons that convert visual representations into language concepts, while \citet{palit2023towards} applied causal tracing to BLIP for visual question answering. Subsequent work extended these methods to generative VLMs: \citet{neo2024vlm} and \citet{yu2024understanding} analyze how LLaVA processes visual information and VQA mechanisms, respectively, \citet{basu_understanding_2024} examines knowledge retrieval, and \citet{jiang2025interpreting} use the logit lens on image representations to detect and edit out hallucinations. At the attention-head level, \citet{basile2025headpursuit} shows that a small subset of heads can be ranked by relevance to semantic or visual concepts and edited to suppress or enhance targeted concepts, \citet{yang-etal-2025-understanding} identify ``hallucination heads'' whose attention patterns mirror the base LLM, and \citet{nikankin2025same} use circuit discovery to show that visual and textual tasks recruit largely disjoint computational subgraphs. Despite these advances, the mechanistic investigation of how VLMs resolve conflicts between modalities remains underexplored.

\paragraph{Internal dynamics of multimodal conflicts.}
Although interest in VLM interpretability is growing, mechanistic studies of how these models process conflicting information remain limited. 
In the context of LLMs, research has focused on understanding how models resolve conflicts between contextual and internal knowledge \citep{ortu2024competition,yu2023characterizing, jin-etal-2024-cutting}. Recent work has begun exploring internal mechanisms in VLMs: \citet{golovanevsky-etal-2025-vlms} study attention heads in LLaVA and BLIP through semantically corrupted image pairs, \citet{Hua2025Conflicting} analyze how modality preference under explicit caption--image conflict is 
reflected in internal representations and modulated by specific heads, and \citet{Golovanevsky2025pixelVSpriors} use steering vectors to control the competition between visual input and parametric knowledge on simple visual attributes. Unlike these, we focus on implicit commonsense conflicts and verify that the identified heads are specifically recruited under conflict 
rather than for general visual processing.

\section{Dataset}
\label{sec:dataset_construction}
\subsection{Requirements for Mechanistic Analysis of Multimodal Conflicts}
Mechanistic interpretability of VLMs requires datasets that enable precise analysis of internal information flow. To support this goal, we identified four key requirements for a suitable dataset:
\begin{itemize}
\item \textbf{Controlled conflict induction}: Conflicts between visual input and internal knowledge must be systematically induced and verifiable, enabling causal analysis.
\item \textbf{Token-level precision}: The dataset should allow token-level inspection and interventions, with prompts designed to elicit specific, predictable continuations.
\item \textbf{Commonsense knowledge grounding}: Scenarios must rely on the model’s internal parametric knowledge, providing strong, consistent priors that can be challenged by visual input. Consistent with the type of commonsense violations studied in \texttt{WHOOPS!} \cite{Bitton2023whoops}, we treat commonsense as robust parametric knowledge encoded in the model’s weights, which serves as a uniform internal prior for analysis.
\item \textbf{Topical generality}: To test broad knowledge and contextual understanding, the dataset should cover a wide range of topics rather than narrow or highly specific domains.
\end{itemize}

To meet these requirements, we construct \texttt{\textsc{WHOOPS-AHA!}}, a dataset specifically designed to support mechanistic interpretability techniques for VLMs. To the best of our knowledge, no existing dataset combines these characteristics, making \texttt{\textsc{WHOOPS-AHA!}} a necessary resource for studying controlled knowledge conflicts in multimodal models. Although designed for our experiments, it may also benefit the broader community interested in mechanistic analysis of multimodal conflicts.

\subsection{Dataset Construction}
\texttt{\textsc{WHOOPS-AHA!}} addresses these requirements by building on the  \texttt{WHOOPS!} collection  \citep{Bitton2023whoops}, which features 500 visually implausible, semantically rich scenes annotated with textual descriptions and explanations of their underlying anomalies.
Each example in \texttt{\textsc{WHOOPS-AHA!}} consists of (i) a counterfactual image depicting an unusual scene, (ii) a sentence referring to the image, and (iii) two sets of plausible continuations: ($\tfactset$) reflecting common sense knowledge, and ($\cfactset$) consistent with the counterfactual scene represented in the image. To align with previous work \citep{ortu2024competition}, we refer to predictions consistent with internal commonsense ($\tfactset$) as \textit{factual}, and those driven by the contradictory visual evidence ($\cfactset$) as \textit{counterfactual}.
\paragraph{Construction pipeline.} For each image in \texttt{WHOOPS!}, we use GPT-4o to generate a sentence that references the anomaly, while remaining consistent with commonsense (factual) completion without visual input. 
GPT-4o is also prompted to produce a set of plausible factual tokens $\tfactset$ and visually-grounded counterfactual continuations $\cfactset$.
For instance, given an image representing a wolf howling at the sun (see \cref{fig:cartoon}), the sentence proposed by GPT-4o is \texttt{``The wolf is howling at the''}, $\tfactset$ = \{\texttt{``moon'', ``night'',...}\} $\cfactset$ = \{\texttt{``sun'', ``daylight'', ``morning'',..}\}. Full prompt details are provided in \cref{app:prompt}.

\paragraph{Quality control and validation.}
To ensure dataset quality, we implemented an LLM-as-a-judge approach \citep{zheng2024llmasajudge}, using GPT-4.1 \citep{openai2025gpt4.1} and Gemini-2.5-Flash  \citep{gemini2025}. Models evaluated each completion for grammatical correctness (1–3 scale) and for alignment with common knowledge or visual anomalies (1–5 scale).
Across the dataset,  the average grammatical score was $2.94\pm0.25$ for completions of inner knowledge and $2.93\pm0.28$ for completions aligned with the image. Alignment with knowledge or visual anomalies received a mean score of $4.43\pm0.97$ and $4.69\pm0.92$, respectively. 

To validate this setup, we compared LLM ratings with those of 2 human evaluators on a 20-item subset. Full details, including prompts, scoring instructions, and agreement results, are provided in \cref{app:llm_judge}.

\section{Background and Methods}

    \subsection{Model Architectures}
    A VLM encodes image-text tokens with a visual encoder and text embeddings, propagating the resulting residual stream through layers with attention and MLP blocks. The final output is projected to the vocabulary space. We focus our analysis on the residual stream, attention, MLP blocks, and individual attention heads.   
    We focus on two models: LLaVA-NeXT-7B \citep{liu2024llavanext} and Gemma3-12B \citep{kamath2025gemma3}. LLaVA-NeXT has 32 layers with 32 attention heads per layer, while Gemma3 has 48 layers with 16 attention heads per layer. Both models use a visual encoder to process image features, but generate only textual output.

\subsection{Analytical Tools}
\label{subsec:analytical_tools} 
\paragraph{Logit inspection.}
To identify the internal components of VLMs responsible for the competition between inner knowledge and conflicting visual context, we apply the \textit{Logit Lens} technique \citep{logitlens2020lesswrong}, which projects intermediate hidden representations into the vocabulary space. This approach has been used in previous work to analyze token-level information flow \citep{nanda2023grokking, halawi2023overthinking_the_truth, yu2023characterizing, ortu2024competition} in LLMs. In our setting, we apply the Logit Lens to the last token of the input and extract the logits corresponding to the tokens in $\tfactset$ and $\cfactset$ in the output of the MLP, Attention block, and across all attention heads, to identify components that favor one mechanism over the other.

\paragraph{Targeted intervention on attention heads.}
    To test the causal role of specific attention heads in promoting predictions aligned with either factual inner knowledge or counterfactual visual context, we intervene on their attention patterns during inference. We define two groups of heads based on Logit Inspection: factual heads ($\mathcal{H}_{\text{fact}}$), which favor predictions based on inner knowledge, and counterfactual heads ($\mathcal{H}_{\text{cofa}}$), which favor visually grounded alternatives. We apply a multiplicative intervention to their attention weights at the final token position (i.e., the last row of the attention matrix), after the softmax operation. 
    Let $\mathbf{A}^{hl}_{\text{last}} = [    \mathbf{A}^{hl}_{\text{last},\text{img}},\mathbf{A}^{hl}_{\text{last},\text{text}}]$ denote the last row of the attention weights for head $h$ at layer $l$, divided between image and text tokens. The intervention is defined as
    \begin{align}
        \mathbf{A}^{hl}_{\text{last}, \text{img}} &\leftarrow (1 - \lambda) \cdot \mathbf{A}^{hl}_{\text{last}, \text{img}}
    \end{align}
if $ (h,l) \in \mathcal{H}_{\text{cofa}}$, and
    \begin{align}
\mathbf{A}^{hl}_{\text{last}, \text{text}} &\leftarrow (1 + \lambda) \cdot \mathbf{A}^{hl}_{\text{last}, \text{text}} 
    \end{align}
    if $(h,l) \in \mathcal{H}_{\text{fact}}$.
        
    This targeted and bidirectional intervention alters the flow of information in a controlled way, allowing us to test whether modulating the influence of these heads changes the model predictions toward the factual or counterfactual outcome. 

    \paragraph{Identification of conflict-inducing visual tokens.}
    To isolate the visual tokens responsible for introducing counterfactual information that conflicts with the inner knowledge of the model, we apply two methods. Both are based on a threshold parameter $\tau \in [0, 1]$, which controls the sensitivity of token selection.
    \begin{enumerate}
        \item \textbf{Most-Attended Visual Tokens:} Given a set of attention heads, we select the visual tokens that receive at least $\tau$ times the maximum attention weight within each head. We then take the union of these tokens across all heads.
        \item \textbf{Gradient-Based Token Importance:} We compute the gradient of the logit associated with a target token (e.g., from $\tfactset$ or $\cfactset$) with respect to the input visual token embeddings. Visual tokens whose gradient magnitudes exceed $\tau$ times the maximum are selected as influential.
    \end{enumerate}
    By varying $\tau$, we control how many image patches are selected, from none when $\tau$ is 1, to all when $\tau$ is 0.
    %
    This allows us to ablate different image portions and analyze how they affect the model predictions.

\paragraph{Quantifying visual attribution.} To evaluate how precisely attribution methods locate counterfactual visual elements, we measure their overlap with ground-truth object regions. We segment the counterfactual object (e.g., the``Amazon'' in the t-shirt) and compute an Attribution Ratio: the average attribution value (attention weight or gradient magnitude) within the segmented mask divided by the average attribution across background patches only. A ratio greater than 1 indicates that the method assigns higher attribution intensity to the counterfactual object than to the background.

\section{Experimental Results}

\subsection{Inducing the Conflict between Inner Knowledge and Visual Context}
\label{sec:whoops}
Given a model, we select $\tfact$ as the highest probability token in $\tfactset$ using text-only prompts, and $\cfact$ as the highest probability token from $\cfactset$ using multimodal input. Selecting from these sets ensures that we capture the completions most aligned with the model’s internal knowledge (text-only) or most influenced by visual information (multimodal), allowing us to reliably study the interaction and potential conflicts between the two sources of information. 
For example, \texttt{``The wolf is howling at the''} yields $\tfact=\text{``moon''}$ (with probabilities of  78\% and 100\% in  LLaVA-NeXT and Gemma3, respectively) in text-only mode, but shift to $\tfact=\text{``sun''}$ (26\% LLaVA-NeXT, 44\% Gemma3) when the image is included, while the probability of \texttt{moon} drops to 17\% and 0.02\%.  After filtering ambiguous cases where counterfactual tokens dominate in text-only scenarios, we retain 436 examples for LLaVA-NeXT and 432 for Gemma3. The systematic shift from factual to counterfactual predictions (factual accuracy drops to 27\% and 24\%, respectively) confirms that visual input successfully overrides internal knowledge.
\begin{figure}[t]
         \centering
         \includegraphics[width=\linewidth]{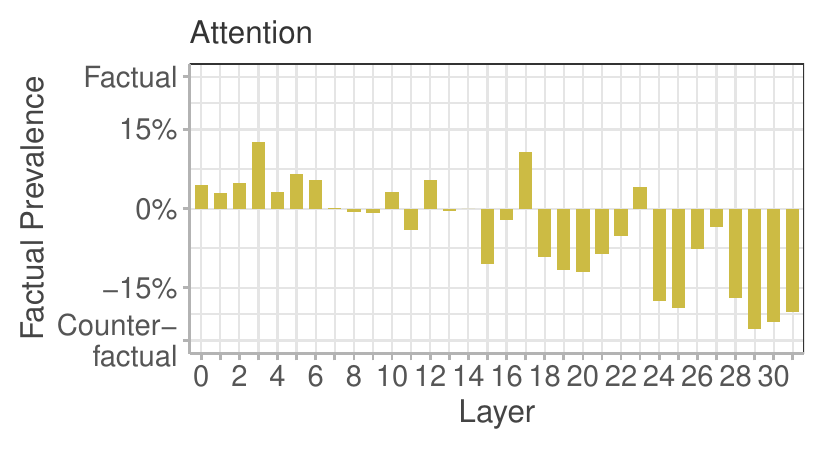}
            \includegraphics[width=\linewidth]{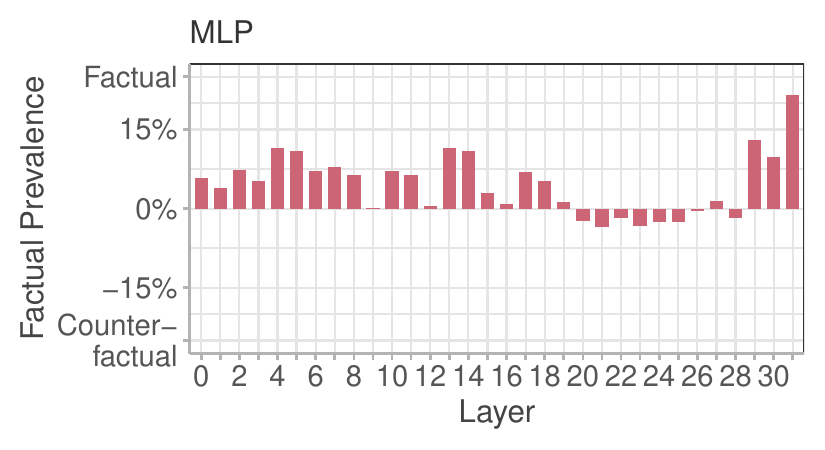}

    \caption{\textbf{Factual prevalence in attention and MLP blocks.} 
    Factual prevalence of LLaVA-NeXT shows whether each block favors predictions aligned with factual knowledge (positive) or counterfactual visual context (negative).    
    The results reveal a functional distinction: attention blocks tend to support counterfactual information (\textbf{top}), whereas MLP blocks frequently promote the model’s internal knowledge (\textbf{bottom}).}
    \vspace{-0.1cm}

    \label{fig:attn_mlp_blocks}
\end{figure}
This setup ensures that the image introduces a counterfactual signal that conflicts with the model’s inner knowledge, allowing us to analyze how visual input alters the model’s prediction compared to its default behavior based on factual knowledge alone.
\begin{figure*}[ht]
     \centering
     \includegraphics[width=\linewidth]{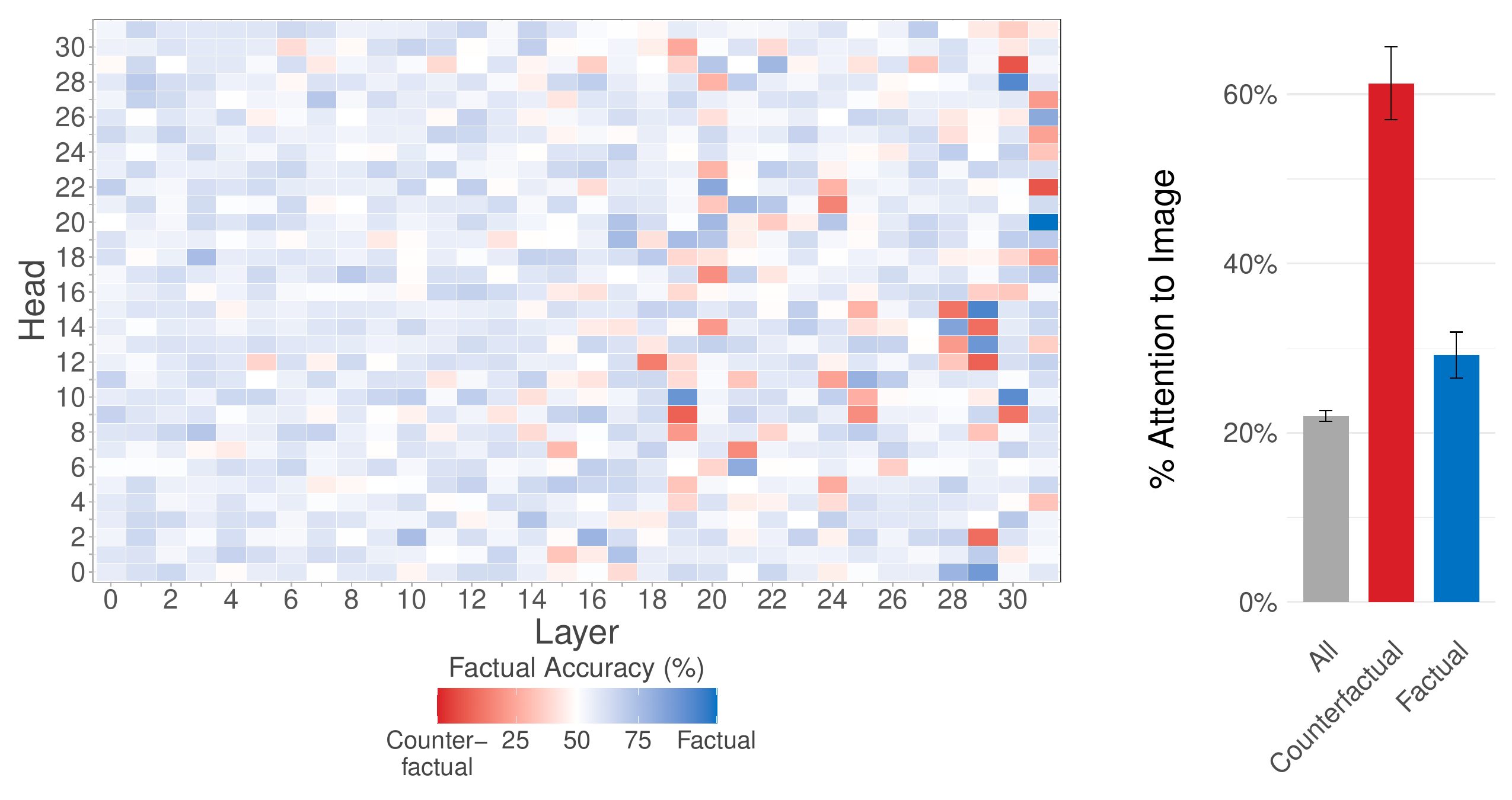}
\caption{\textbf{Contribution of attention heads to factual and counterfactual predictions.} 
(\textbf{Left}) Factual accuracy of individual attention heads in LLaVA-NeXT, based on Logit Lens projections at the final token position. Blue indicates heads that tend to favor the factual token (reflecting inner knowledge), while red indicates heads that favor the counterfactual token (introduced by the visual context). 
(\textbf{Right}) Mean attention to image tokens at the final generation step for heads in each group. Each group contains $20$ attention heads. Counterfactual heads attend significantly more to the image (60\%) than factual heads (28\%) or the model-wide average (22\%), indicating that visual information is directly propagated to the output and plays a key role in counterfactual predictions.}
     \label{fig:loc_heads}

 \end{figure*}

\subsection{The Tension Between Inner Knowledge and Visual Context is Localized}
\label{sec:attention_heads}
Building on the controlled knowledge conflict, we apply Logit Lens to identify which model components mediate the competition between $\tfact$ and $\cfact$. For attention and MLP blocks, we report factual preference strength—the deviation from random baseline (0.5) in factual accuracy—where positive values indicate bias toward internal knowledge and negative values toward visual context. For individual attention heads, we report raw factual accuracy (the fraction of examples where factual logits exceed counterfactual logits) to identify heads with strong directional preferences.

\paragraph{Functional separation between attention and MLP layers.} 
We first compare attention and MLP contributions to predicting $\tfact$ and $\cfact$ (\cref{fig:attn_mlp_blocks} for LLaVA-NeXT; see \cref{app:gemma} for Gemma3).
Attention blocks exhibit a stronger tendency to favor the counterfactual visual context, whereas MLP blocks are more aligned with the internal factual knowledge. In particular, the influence of attention blocks increases from the middle layers (around layer 15), peaking in the final four layers. 
MLP blocks similarly show their strongest alignment to factual knowledge in the upper layers, with a peak at the final layer, consistent with prior findings on upper-layer MLPs retrieving factual knowledge \citep{geva2021keyvalue,meng2022rome,dai2021knowledgneneurons}.

\paragraph{Localization of the modality conflict to individual attention heads.}
We next examine the role of individual attention heads. \Cref{fig:loc_heads}-left shows the tendency for each attention head to promote or suppress the factual token in LlaVa-NeXT (see \cref{fig:gemma_loc_heads} for Gemma3).
The distribution shows that only a small subset of heads exhibit a strong, consistent alignment with $\tfact$ or $\cfact$. 
Moreover, consistent with the results at the block level, these factual and counterfactual heads are concentrated in the final layers of the model, indicating that the conflict between inner knowledge and visual context is resolved late in the forward pass.   
In the analyses that follow, we focus on the 20 attention heads that most strongly promote the factual and counterfactual tokens. We chose 20 heads as this provides an optimal balance: it maximizes factual accuracy while minimizing potential disruptions to model stability that could arise from intervening in too many heads (see \cref{app:n_heads}). On average, the factual heads favor the $\tfact$ 85\% of the time, and the counterfactual ones $\cfact$ 15\% of the time, indicating strong alignment with their respective information sources.

\paragraph{Factual and counterfactual heads exhibit distinct visual attention patterns.}    
We then investigate whether heads associated with the factual mechanism or the counterfactual visual context exhibit distinct attention patterns -- specifically, whether they attend to different token modalities (image or text). 
Since the counterfactual information is introduced through the image, a natural hypothesis is that counterfactual heads attend more strongly to visual tokens, while factual heads rely more on textual content. 
To test this hypothesis, for each group of heads, we sum the attention weights assigned to visual tokens in the last row of each head and average across the dataset. 
\Cref{fig:loc_heads}-right reports the average amount of attention to the image for the two groups of heads. 
Heads favoring the counterfactual token $\cfact$ attend to image tokens significantly more (61\%) than those aligned with inner knowledge (29\%) or the model-wide average (22\%). 
Although the counterfactual signal originates in the image, it is not a priori clear that this information is transmitted directly to the final token. 
%
The model could, in principle, diffuse or encode this signal in different positions across intermediate layers. 
However, the observed attention patterns suggest that the visual context influences the output token directly in late layers of the model, with limited intermediate processing. 
These findings are consistent for Gemma3 (see \cref{app:gemma}).
    
\subsection{Targeted Intervention on Selected Attention Heads Causally Shifts Model Behavior}
\label{sec:intervention}
Having identified attention heads aligned with either factual knowledge or counterfactual visual context, we next examine whether these components play a causal role in shaping model predictions. 
To this end, guided by our earlier observation that counterfactual heads attend more to visual tokens, we apply the targeted bidirectional intervention strategy described in \cref{subsec:analytical_tools} that selectively adjusts attention values based on head type and token modality, modifying the attention weights to steer the output of the model towards one mechanism or the other.  As a control experiment to isolate the effect of targeted interventions, we randomly select 100 attention heads and apply the same intervention for varying $\lambda$ values. 
This manipulation does not produce a substantial deviation from the baseline.
The complete results for the control experiment are reported in \cref{app:n_heads}.

\begin{figure}[t]
\centering
\includegraphics[width=\linewidth]{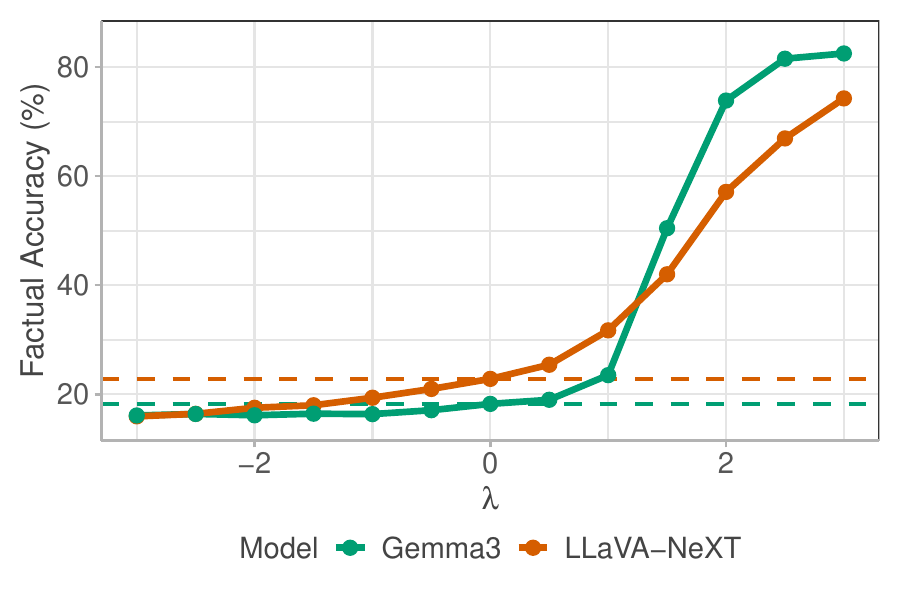}
\caption{\textbf{Intervention on target attention heads.} Change in factual accuracy under different levels of intervention strength ($\lambda$). For $\lambda < 0$, we boost the counterfactual heads (on image tokens) and weaken the factual heads (on text tokens); for $\lambda > 0$, we do the opposite. The intervention is applied at the final token position, modifying only the relevant attention values in the last row.}
\label{fig:heads_intervention}
\vspace{-0.4pt}
\end{figure}

\Cref{fig:heads_intervention} shows the results of our intervention for LLaVA-NeXT (orange profile) and Gemma3 (green profile). 
When we increase attention from factual heads and decrease it from counterfactual heads using LLaVA-NeXT, the factual accuracy increases to 74\%, indicating a strong shift towards predictions of inner knowledge. 
Conversely, reversing the intervention reduces the accuracy to 16\%, confirming that these heads causally influence whether the model favors factual or counterfactual content.
A similar trend can be observed for Gemma3, with an even stronger relative effect driven by its lower baseline factual accuracy of 18\% and a peak of $83\%$. 
Comprehensive details about the choice of the parameter $\lambda$ are reported in \cref{app:lambda_selection}. 
To verify that $\mathcal{H}_\text{cofa}$ are not generic
vision-centric heads, we evaluate them on \texttt{\textsc{POPE}}~\citep{li2023pope},
a standard binary VQA benchmark designed to probe general visual
grounding. Suppressing their image attention leaves accuracy
completely unchanged in both models, while removing the image
entirely collapses performance to chance. This dissociation confirms
that the identified heads are not required for routine visual
processing, but are selectively recruited when parametric knowledge
and visual evidence conflict. Full results are in
Appendix~\ref{app:pope}.
To assess whether the identified mechanisms generalize beyond \texttt{\textsc{WHOOPS-AHA!}}, we replicate the same head identification and intervention procedure on an independently constructed \texttt{visual counterfactual} dataset with more photo-realistic images. The results closely mirror our main findings and are reported in \cref{app:generalization_vcf}.

\subsection{Counterfactual Predictions Depend on Localized Image Regions}
\label{sec:attention_attribution}

The previous analysis reveals that specific attention heads at the final token position mediate the conflict between contextual information and internal knowledge, with heads aligned with the visual context strongly attending to image tokens, injecting visually grounded information into the generation process.
\begin{figure}[t]
\centering
\includegraphics[width=\linewidth]{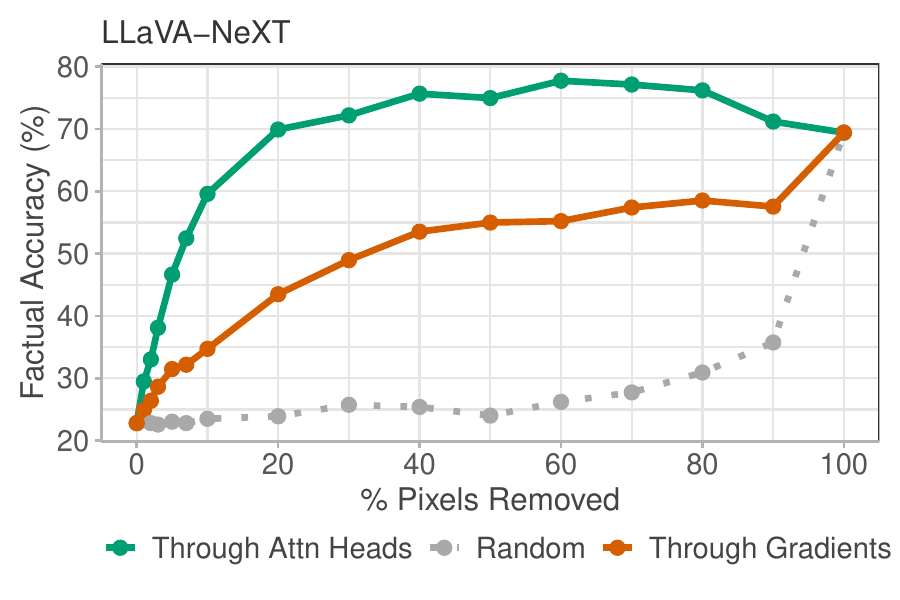}
\includegraphics[width=\linewidth]{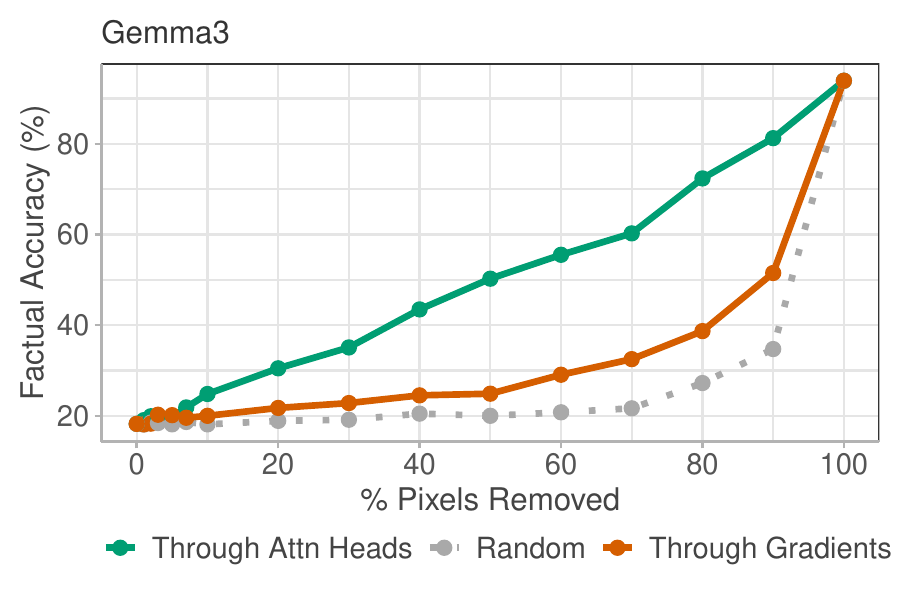}
\caption{\textbf{Ablation of relevant pixels.} The plot shows the effect of ablating different percentages of image pixels in LLaVA-NeXT. The green line corresponds to pixels selected based on the highest attention from counterfactual heads, while the orange line corresponds to pixels with the highest gradient magnitude with respect to the counterfactual token. The gray line shows a random baseline where pixels are removed uniformly at random.}
\label{fig:pixel_intervention}
    \vspace{-0.3cm}

\end{figure}
However, two key questions remain open. (i) Is the counterfactual visual signal localized to specific image regions or spread across the input? 
(ii) Is the visual signal passed directly to the last token position, or is it mediated by successive layers and tokens before reaching the output in the upper layers?
\begin{figure}[t]
\centering
\includegraphics[width=\linewidth]{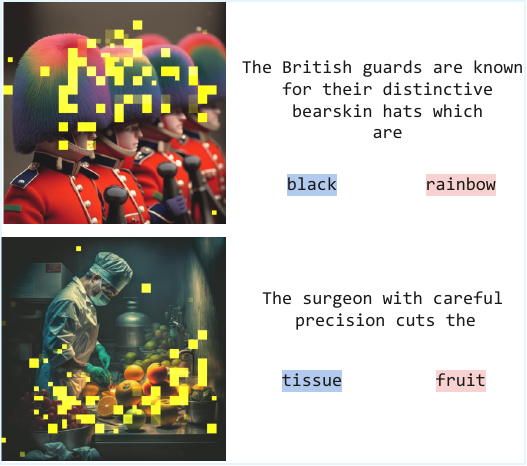}
\caption{\textbf{Visual regions driving counterfactual predictions.} Highlighted image regions, identified through attention-based attribution, show the most influential visual patches for counterfactual predictions.
The highlighted areas align with semantically meaningful, visually anomalous content, indicating that counterfactual outputs are grounded in localized image features.}
\label{fig:qualitative_examples}
\vspace{-0.3cm}
\end{figure}
To address these, we conduct two analyses: (i) using attention and gradient-based attribution to identify the image patches driving counterfactual predictions, as described in \cref{subsec:analytical_tools}; and (ii) ablating these patches by setting their visual token embeddings to zero and measuring the change in factual accuracy. A control experiment is also performed where an equivalent number of randomly selected patches are ablated.

\paragraph{Patch attribution and ablation effects.}
The results (\cref{fig:pixel_intervention}) show that ablating patches identified through attention-based attribution leads to a sharp and consistent increase in factual accuracy as more pixels are removed (green profiles). 
For LLaVA-NeXT, factual accuracy improves markedly with the ablation of just 10–30\% of the top-ranked patches and eventually plateaus around 80\%. 
Gradient-based attribution (shown in red) also yields a substantial increase in factual accuracy, but with less pronounced effects, suggesting lower precision in identifying counterfactual-driving regions. 
In contrast, ablating an equivalent number of randomly selected patches results in only minor fluctuations in accuracy. These findings confirm the causal role of the identified regions and support the hypothesis that counterfactual signals are spatially localized and semantically specific.

\paragraph{Localization of counterfactual visual evidence.}
To assess the semantic coherence of the identified visual regions, we qualitatively inspect examples where attribution highlights patches responsible for counterfactual predictions (\cref{fig:qualitative_examples}). In many cases, these patches correspond to intuitive visual elements that directly contradict commonsense expectations, such as implausible objects or substitutions. For example, when the model predicts “rainbow” instead of “black” for a bearskin hat, the highlighted regions focus on the hat’s unrealistic coloring, and when “fruit” replaces “tissue” in a surgical scene, attention concentrates on the unexpected oranges on the operating table.

We complement this analysis with a quantitative experiment on 20 randomly selected images, using SAM-3 \citep{sam3} to segment the counterfactual objects. We compare attention from the identified counterfactual heads to both gradient-based attribution and a random-head baseline. Using the \textit{Average Ratio} metric (\cref{subsec:analytical_tools}), we find that for Gemma3 the counterfactual heads strongly concentrate attention on the segmented regions (median $4.41$, IQR $1.09$–$7.85$), significantly outperforming both gradients (median $1.74$) and random heads (median $0.92$).  These results quantitatively validate that the heads identified by our method act as precise pointers to the visual evidence driving the counterfactual override (see \cref{app:segmentation} for full details).

\section{Discussion}

The competition between parametric knowledge and conflicting external evidence, identified in text-only LLMs~\citep{ortu2024competition, jin-etal-2024-cutting, wu2024retrieval, yu2023characterizing}, extends to the multimodal setting: parametric--visual conflicts are resolved by a small, identifiable subset of attention heads in the upper layers of the model. The functional dissociation between attention and MLP blocks, with attention favoring visual context and MLPs reinforcing parametric priors, aligns with the established view that upper-layer MLPs store factual associations~\citep{geva2023dissecting, meng2022rome} while attention heads integrate contextual signals. Subsequent work has converged on related findings: \citet{Golovanevsky2025pixelVSpriors} trace the layer-wise evolution of modality preference via early decoding and propose steering vectors to control it, while \citet{Hua2025Conflicting} show that specific attention heads modulate modality preference under explicit caption--image conflict. Relative to both, our setting is distinctive in studying implicit commonsense conflicts rather than explicit modality instructions or simple visual attributes, and in establishing that the identified heads are specifically recruited under conflict rather than being generic vision-processing components. \citet{golovanevsky-etal-2025-vlms} similarly identify functionally specialized heads in LLaVA and BLIP, though without focusing on conflict resolution. 

The additional finding that conflict-resolution heads encode \textit{where} in the image the contradicting evidence originates suggests that mechanistically identified circuits can serve as a byproduct attribution tool, complementing approaches that use internal representations for visual 
grounding~\citep{jiang2025interpreting, phukan2024beyond} and functionally specialized heads~\citep{yang2025nullu, sarkar-etal-2025-spin, he2024cracking}.

Our analysis is bounded by its focus on late-fusion architectures and the logit lens, which introduces known approximation errors~\citep{belrose2023tunedlens}; extending this framework 
to free-form generation, earlier-fusion models, and broader conflict types remains important future work.

\section{Conclusion}

We introduced \texttt{\textsc{WHOOPS-AHA!}} and a mechanistic pipeline to study how VLMs resolve conflicts between visual evidence and internal parametric knowledge. A small set of upper-layer attention heads mediates this competition with clear functional specialization: counterfactual heads amplify visual evidence while factual heads reinforce parametric priors, 
and targeted interventions on them causally shift model behavior in both directions. These heads are conflict-specific rather than generically visual, and their attention patterns localize counterfactual image regions more precisely than gradient-based attribution. We hope these findings 
provide a foundation for building multimodal systems that more reliably calibrate reliance on vision and knowledge.

\section*{Limitations}
\paragraph{Methodological limitations.} The analysis relies on the Logit Lens technique to project intermediate hidden states into token logits. Although this method has been widely adopted for interpretability, it is known to introduce distortions due to projection from non-final residual states \citep{belrose2023tunedlens}, and should be interpreted as an approximate diagnostic rather than a precise decoding proxy. In our setting, we use a representative factual and counterfactual token per example to enable controlled comparisons. Although this simplifies the generative landscape of the model, it offers a practical and interpretable probe of the underlying mechanisms. Future work could explore more model behavior across full generations to complement this approach. Our attribution and intervention methods focus on attention heads and target the final token position. This design isolates interpretable causal signals while remaining tractable, though it does not capture the possible contributions of other components, such as MLP layers or visual encoders. Extending this framework to broader architectural elements is a promising direction.
\paragraph{Scope and generalizability.} We focus on late-fusion, LLaVA-style architectures, which are particularly well-suited for controlled image-understanding tasks. These models are among the best open-source architectures for visual understanding, making them ideal for the interpretability methods employed in our study. Our interest is specifically in how visual input interacts with internal knowledge during textual generation, so we chose models that are designed with a focus on image understanding. While early or mid-fusion models also use attention to integrate visual features into the language stream, they may differ significantly in how information is communicated between the modalities \citep{Serra2025narrowgate}. The point of injection of visual features varies, but the underlying mechanism of cross-modal communication through attention remains consistent across these models. By focusing on late-fusion models, we ensure a more controlled and traceable examination of visual-to-text interactions in widely used open source multimodal models, though this choice limits the generalizability of our findings to models with different fusion strategies.

\section*{Ethical Considerations}
This work aims to improve our understanding of how VLMs resolve conflicts between internal factual knowledge and contradictory visual context. Our analysis is intended to contribute to foundational research in model interpretability, with the broader goal of developing more transparent and controllable multimodal systems. The techniques presented are diagnostic and exploratory in nature, designed to support responsible development and analysis of multimodal systems. We believe that studying the dynamics of conflicting information sources is essential for anticipating model failure modes, mitigating unintended behaviors, and building more robust AI systems. All models and data are used in accordance with their intended research licenses, and \texttt{\textsc{WHOOPS-AHA!}} is released solely for non-commercial, research purposes under compatible terms. We used AI assistants (e.g., GitHub Copilot) to support code completion during experiment implementation; all generated code was manually reviewed and supervised by the authors.

\section*{Acknowledgments}
The authors acknowledge the AREA Science Park supercomputing platform ORFEO made available
for conducting the research reported in this paper and the technical support of the Laboratory of Data Engineering staff.
Francesco Ortu, Diego Doimo and Alberto Cazzaniga were supported by the project ``Supporto alla diagnosi di malattie rare tramite l’intelligenza artificiale" CUP: F53C22001770002 and ``Valutazione automatica delle immagini diagnostiche tramite l’intelligenza artificiale", CUP: F53C22001780002. A. C. acknowledges financial support under the National Recovery and Resilience Plan (NRRP), mission 4, component 2, investment 1.1, and call for tender no. 1409 published on 14 September 2022 by the Italian Ministry of University and Research (MUR), funded by the European Union — NextGenerationEU — CUP J53D23015070001.

{
    \small
       \bibliography{references}
}

\clearpage
\appendix
\section{Reproducibility}
We ran the experiments on one NVIDIA H100 GPU, and two GPUs for the gradient-based attribution tests. 
We use the HuggingFace Transformers library \citep{wolf-etal-2020-transformers} with public implementations of LLaVA-NeXT and Gemma3. The total compute time is 20 GPU hours. 
The \texttt{WHOOPS!} dataset was released with a CC-By 4.0 license.
\section{LLM-as-a-Judge: Detailed Validation and Analysis}
\label{app:llm_judge}

\subsection{Evaluation Setup}
We used GPT-4.1 (gpt-4.1-2025-04-14) and Gemini-2.5-Flash (gemini-2.5-flash-image-preview) through OpenRouter \citep{openrouter2025} to evaluate each dataset completion along two dimensions:
\begin{itemize}
\item Grammatical correctness:
\begin{itemize}
    \item 1 (No) All completed sentences contain grammatical errors.
    \item 2 (Some do not make sense) Some completed sentences have grammatical errors.
    \item 3 (Yes) All completed sentences follow proper grammar rules.
\end{itemize}

    \item Knowledge/Anomaly Alignment (1–5 scale)
        \begin{itemize}
            \item 1 = Poor alignment: Completion ignores or misrepresents common knowledge or visual anomalies
            \item 5 = Excellent alignment: Completion clearly reflects correct knowledge or accurately captures anomalies in the image.
        \end{itemize}
\end{itemize}
\subsection{Aggregate Statistics}

\begin{table*}[ht]
\centering
\resizebox{\textwidth}{!}{
\begin{tabular}{lccccc}
\toprule
\textbf{Metric} & \textbf{Mode} & \textbf{Gemini-2.5-Flash} & \textbf{GPT-4.1} & \textbf{Average} & \textbf{Exact Agreement} \\
\midrule
\multirow{2}{*}{Grammatical Correctness} & Text-only & $2.95 \pm 0.23$ & $2.93 \pm 0.26$ & $2.94\pm0.25$ & $95.0\%$ \\
 & With Image & $2.92 \pm 0.30$ & $2.93\pm0.26$ & $2.93\pm0.28$  & $92.8\%$\\
\midrule
\multirow{2}{*}{Knowledge/Anomaly Alignment} & Text-only & $4.55\pm0.98$ & $4.31\pm0.94$ & $4.43\pm0.97$ & $69.5\%$ \\
 & With Image & $4.76\pm0.93$ & $4.60\pm0.91$ & $4.68\pm0.92$ & $80.2\%$\\
\bottomrule
\end{tabular}
}
\caption{\textbf{LLM-as-a-judge evaluation results.} Gemini-2.5-Flash and GPT-4.1 for Text-only and Image-based Scenarios}
\label{app_tab:llm_as_a_judge}
\end{table*}

We report the average results for all dimensions in \cref{app_tab:llm_as_a_judge}. The results show that the judges rated the sentences and the completions as mostly grammatically correct and aligned with the common knowledge or the anomaly in the image.

To measure agreement between the two judges, we calculate the exact agreement rate. We observed a strong ceiling effect in our data, with the distribution of responses being heavily imbalanced. Overall, the results demonstrate that the LLM as a judge setup produces consistent and interpretable ratings across both factual and counterfactual scenarios, with high alignment between models (Gemini vs. GPT-4.1). These findings support the reliability of LLM-based evaluation for large-scale dataset validation.

\subsection{Human Validation}
To assess how well the LLM-as-a-judge setup reflects human evaluation, we sampled 20 entries from our dataset and asked 2 human evaluators, both graduate students at a local institution, to complete the same tasks as the LLMs. To measure agreement between human evaluators and LLM judgments, we averaged the ratings of the human evaluators and rounded to the nearest integer for each category, and then measured the exact and adjacent agreement rates.

For grammatical correctness in text-based completions, we observed 60\% agreement, and 70\% for image-associated completions. For alignment, the agreement was 35\% for text-only completions and 65\% for image-associated completions. However, when considering adjacent category agreement (i.e., when responses were within one point of each other on the scale), the agreement increased to 85\% for text-based completions and 100\% for image-associated completions.

\subsection{Prompt and Instruction}

\begin{tcolorbox}[mycustombox={Prompt Used to Instruct LLM-as-a-Judge with Image}]

You will evaluate sentence completions that should describe UNUSUAL or ANOMALOUS elements visible in the provided image. Focus on identifying completions that capture strange, unexpected, or contradictory visual elements.

    EVALUATION CRITERIA:

    1. GRAMMATICAL CORRECTNESS: Are the completed sentences grammatically valid?
    
    Check syntax, word order, agreement, and grammatical structure of each completed sentence.
    
    → "Yes" = ALL completed sentences follow proper grammar rules
    → "Some do not make sense" = SOME completed sentences have grammatical errors
    → "No" = ALL completed sentences contain grammatical errors
    
    Focus on: subject-verb agreement, article usage, word order, tense consistency

    2. KNOWLEDGE ALIGNMENT: How well do completions reflect unusual/anomalous visual elements in the image?
    IMAGE CONTEXT: Rate how accurately completions describe the strange/unusual elements you can SEE in the image.
    
    SCORING RUBRIC:
    • 5 = Perfectly captures unusual/anomalous visual elements in the image - highly accurate and relevant
    • 4 = Mostly accurate - minor gaps or imprecisions  
    • 3 = Partially accurate - some correct elements, some missing/wrong
    • 2 = Minimally accurate - mostly incorrect with few relevant elements
    • 1 = Completely inaccurate - fails to reflect unusual/anomalous visual elements in the image

    SENTENCE: "\{text\}"
    COMPLETIONS TO EVALUATE: \{tokens\}

   INSTRUCTION: Look carefully at the image and identify what appears unusual, unexpected, or anomalous. Then evaluate how well the completions capture these strange elements.

    IMPORTANT: You must respond with valid JSON only. Do not include any text before or after the JSON.
    
    Required JSON format:
    \{\{
        "grammatically\_correct": "Yes, all are grammatically correct" | "Some are not grammatically correct" | "No, they are not grammatically correct", 
        "knowledge\_reflection\_score": 1-5,
        "reasoning": "Explain your evaluation for each criterion, citing specific completions"
    \}\}"""
\end{tcolorbox}

\begin{tcolorbox}[mycustombox={Prompt Used to Instruct LLM-as-a-Judge without image }]

 You will evaluate sentence completions that should reflect normal, expected real-world knowledge real-world scenarios based on common knowledge and typical experiences.

    EVALUATION CRITERIA:

    1. GRAMMATICAL CORRECTNESS: Are the completed sentences grammatically valid?
    
    Check syntax, word order, agreement, and grammatical structure of each completed sentence.
    
    → "Yes" = ALL completed sentences follow proper grammar rules
    → "Some do not make sense" = SOME completed sentences have grammatical errors
    → "No" = ALL completed sentences contain grammatical errors
    
    Focus on: subject-verb agreement, article usage, word order, tense consistency

    2. KNOWLEDGE ALIGNMENT: How well do completions reflect "normal, expected real-world knowledge?
    REAL-WORLD CONTEXT: Rate how well completions reflect typical, widely-accepted real-world scenarios.
    
    SCORING RUBRIC:
    • 5 = Perfectly captures normal, expected real-world knowledge - highly accurate and relevant
    • 4 = Mostly accurate - minor gaps or imprecisions  
    • 3 = Partially accurate - some correct elements, some missing/wrong
    • 2 = Minimally accurate - mostly incorrect with few relevant elements
    • 1 = Completely inaccurate - fails to reflect normal, expected real-world knowledge

    SENTENCE: "\{text\}"
    COMPLETIONS TO EVALUATE: \{tokens\}

   Consider what would be normal, expected, and typical in real-world scenarios. Then evaluate how well the completions reflect this common knowledge.

    IMPORTANT: You must respond with valid JSON only. Do not include any text before or after the JSON.
    
    Required JSON format:
    \{\{
        "grammatically\_correct": "Yes, all are grammatically correct" | "Some are not grammatically correct" | "No, they are not grammatically correct", 
        "knowledge\_reflection\_score": 1-5,
        "reasoning": "Explain your evaluation for each criterion, citing specific completions"
    \}\}"""
\end{tcolorbox}
\begin{tcolorbox}[mycustombox={Instruction Given to Human Evaluator}]
\textbf{Task Instructions}
You will be asked to evaluate sentences and their possible completions. Sometimes, an image will also be provided. Your job is to judge whether the completions are appropriate, whether the sentence is grammatically correct, and how well the sentence and completions reflect knowledge or the content of the image.

Please follow these criteria carefully for each question:

1. Is the sentence grammatically correct? \\
Ignore meaning here; only focus on grammar and syntax.
Mark "Yes" if the base sentence with all the possible completions is grammatically well-formed.
Mark "Some do not make sense" if at least one completion create a grammatically incorrect sentence.
Mark "No" if the sentence with all the possible completions contains grammar errors.
2. How well do the sentence and completions reflect common knowledge (or reflect strange/anomalous things in the image)?\\
If the question refers to common knowledge: 
Judge how typical, reasonable, or widely accepted the sentence + completion is.\\
Example: \textit{“The sun rises in the east”} should score high (5).\\
Example: \textit{“The sun rises in the north”} should score low (1).\\
If the question refers to strange/anomalous things in the image:
Judge how accurately the sentence and completions capture unusual, odd, or unexpected elements visible in the image.
Score higher if the completion clearly reflects what is strange in the image. \\
Score lower if it ignores or misrepresents the anomaly.
Use the 1–5 scale consistently: \\

1 = Not at all accurate/appropriate\\
3 = Neutral or partially accurate\\
5 = Very accurate and appropriate\\
Please read each question carefully and provide your evaluations with attention.
\end{tcolorbox}

\section{MLP Intervention}
\label{app:mlp_intervention}
\begin{figure}[h]
\centering
\includegraphics[width=\linewidth]{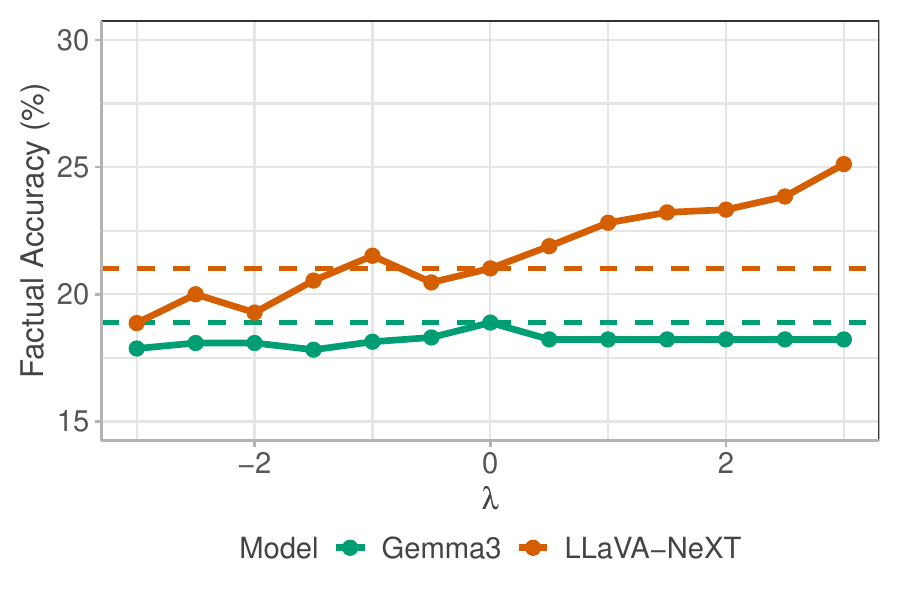}
\caption{\textbf{Effect of MLP interventions.}
Factual accuracy when intervening on the last three MLP blocks at the final token position in LLaVA-NeXT and Gemma3. The observed improvements are minor compared to targeted attention-head interventions (\cref{fig:heads_intervention}).}
\label{app:fig_mlp_intervention}
\end{figure}

We tested whether intervening on MLP blocks could produce effects comparable to those observed with attention heads. Specifically, we applied interventions to the last three MLP blocks at the final token position in both LLaVA-NeXT and Gemma3. The results, reported in \cref{app:fig_mlp_intervention}, show only marginal changes in factual accuracy relative to the baseline. This effect is substantially weaker than the gains obtained from targeted attention-head interventions (\cref{fig:heads_intervention}).

These findings reinforce two key conclusions. First, MLP interventions are less precise: they modify the residual stream broadly, affecting a much larger number of parameters and acting indiscriminately across modalities. This broad influence makes it harder to isolate causal mechanisms and increases the risk of introducing unintended side effects. Second, the limited efficacy of MLP interventions indicates that factual–counterfactual conflicts are primarily mediated by attention mechanisms, not by MLP transformations. This aligns with prior evidence that late-layer MLPs often retrieve factual associations, whereas attention heads are more directly responsible for integrating conflicting cross-modal signals. For these reasons, our analysis centers on attention interventions, which provide both stronger causal leverage and more interpretable control over the balance between internal knowledge and visual input.

\section{Experimental Analysis for Gemma-12b}
\label{app:gemma}

\begin{figure}[h]
\centering
\includegraphics[width=\linewidth]{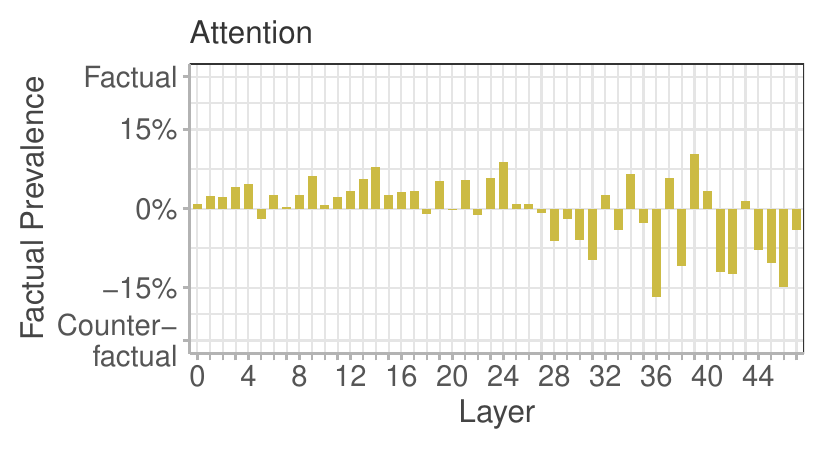}
\includegraphics[width=\linewidth]{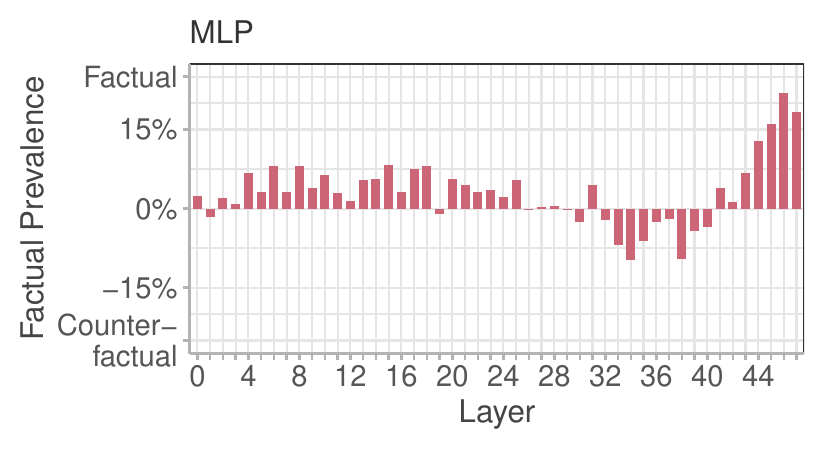}
\caption{\textbf{Factual and counterfactual contributions of MLP and attention blocks in Gemma3.} 
Layer-wise deviation from 50\% factual accuracy for attention and MLP blocks, as measured by the relative logits of $\tfact$ and $\cfact$ via Logit Lens. Positive values indicate a bias toward the factual token, while negative values indicate preference for the counterfactual token. Consistent with trends observed in LLaVA-NeXT, attention blocks in Gemma3 increasingly support counterfactual predictions in higher layers, while MLP blocks show stronger alignment with internal factual knowledge.}
\end{figure}

\begin{figure*}[h]
 \centering
 \includegraphics[width=\linewidth]{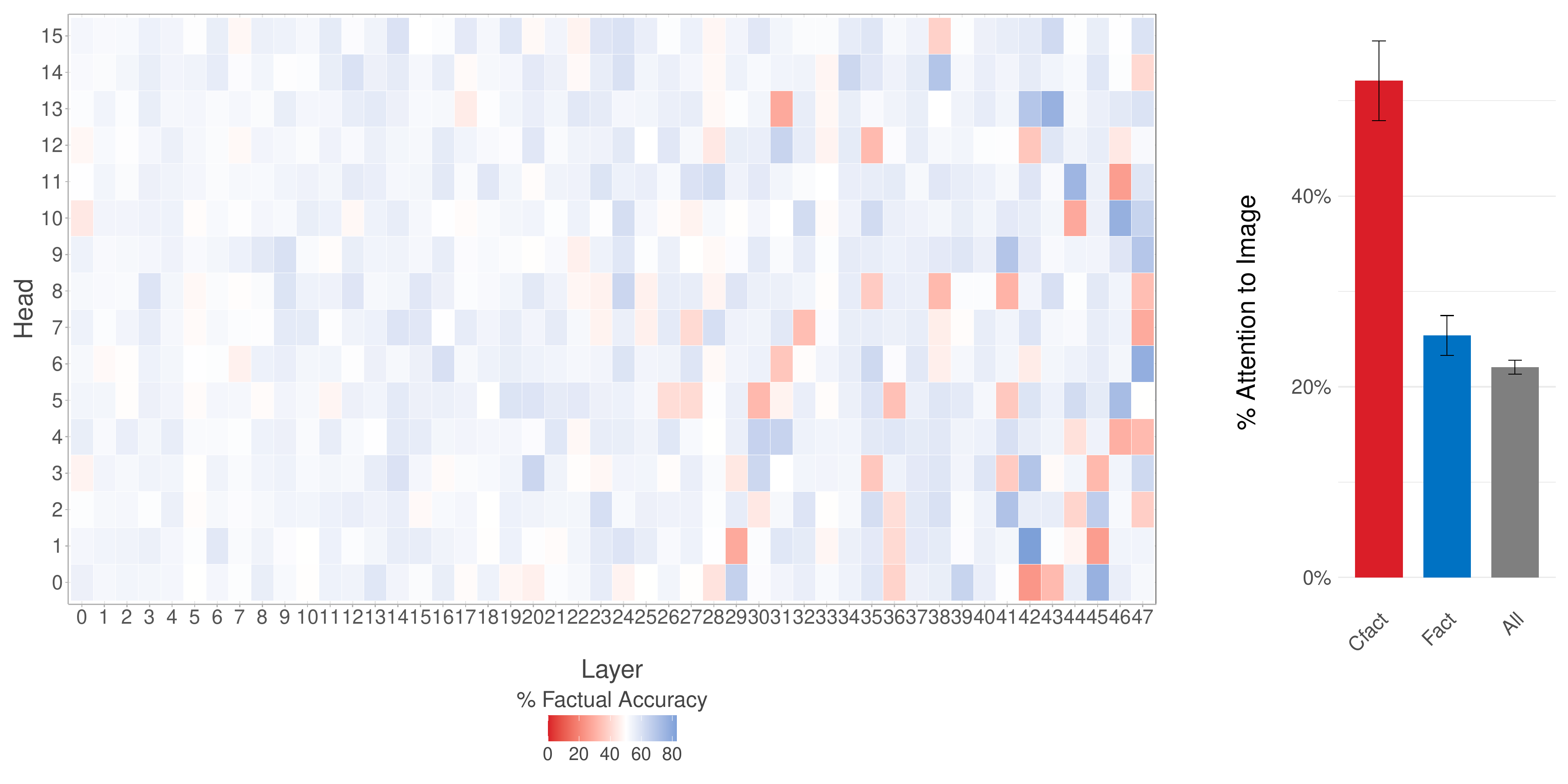}
 \caption{\textbf{Factual and counterfactual contributions of attention heads for Gemma3.} 
(\textbf{Left}) Factual accuracy of individual attention heads in Gemma3, computed using Logit Lens projections of the final token's hidden state. Blue indicates heads that more frequently favor the factual token ($\tfact$), while red indicates those that favor the counterfactual token ($\cfact$). As in LLaVA-NeXT, highly polarized heads are concentrated in the upper layers. 
(\textbf{Right}) Mean attention to image tokens at the final generation step. Counterfactual heads attend more strongly to image tokens (52\%) than factual heads (25\%) or the model-wide average (22\%), highlighting the direct role of visual input in modulating counterfactual predictions.}
\label{fig:gemma_loc_heads}
\end{figure*}

\begin{table*}[h]
\centering
\renewcommand{\arraystretch}{1.3}
\begin{tabularx}{\textwidth}{
  >{\centering\arraybackslash}m{0.22\textwidth} | 
  >{\centering\arraybackslash}m{0.06\textwidth} | 
  X
}
\toprule
\textbf{Image} & \textbf{$\lambda$} & \textbf{Caption} \\
\hline
\multirow{3}{*}{
  \begin{minipage}[c][4.5cm][c]{0.22\textwidth}
    \centering
    \includegraphics[width=0.9\linewidth]{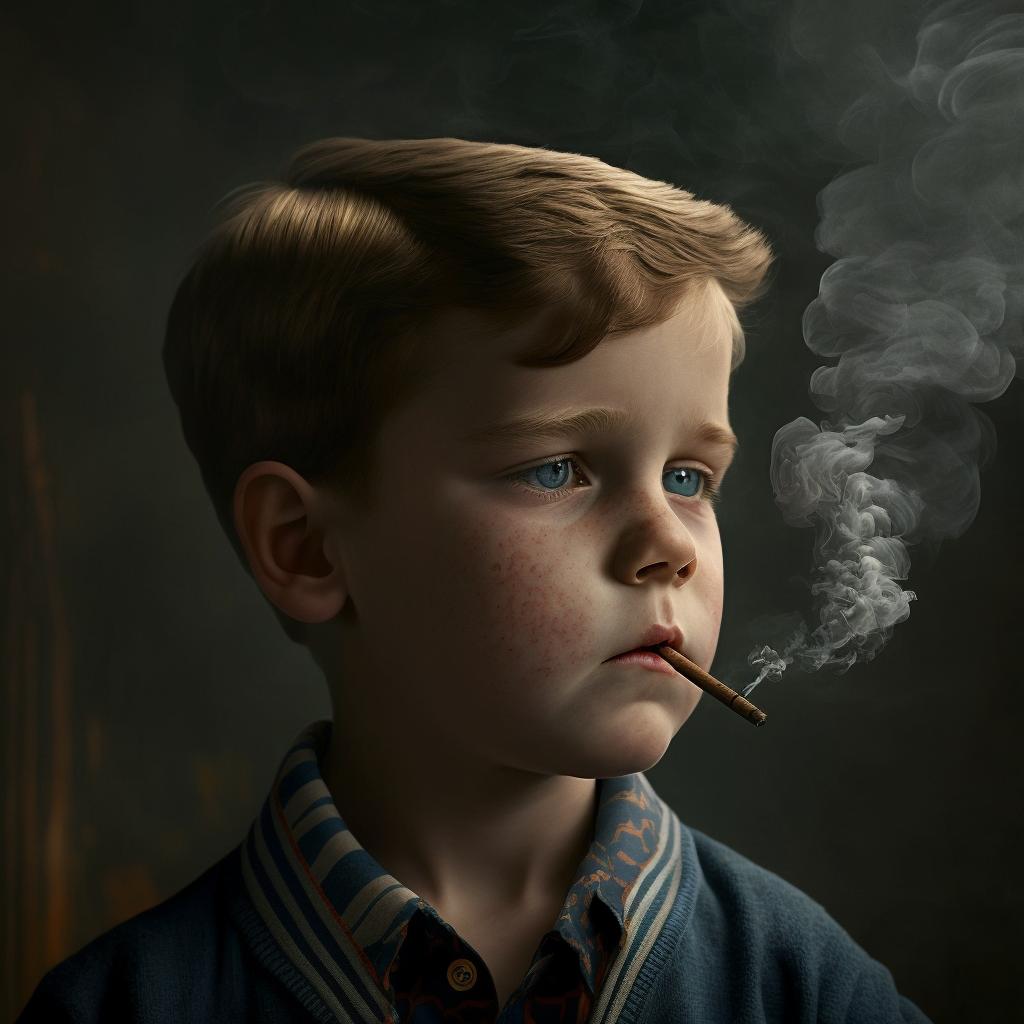}
  \end{minipage}
}
    & $0$   & {\footnotesize The image is a digital artwork of a young boy with a contemplative expression. He has short, light brown hair and striking blue eyes. The boy is wearing a striped shirt with a collar and a patterned tie. } \\
    & $3$  & {\footnotesize The image is a digital artwork of a young child. The child is depicted with a contemplative expression, looking slightly to the side with a thoughtful gaze. They are holding a piece of paper or a small object in their hand, which appears to...} \\
    & $10$ & {\footnotesize Jimmy Wooster spr spr spr spr spr spr spr spr spr spr spr spr spr spr spr spr spr spr spr spr spr spr spr spr spr spr spr spr spr spr spr spr spr spr spr spr spr spr spr spr spr spr spr spr spr spr} \\
\hline
\multirow{3}{*}{
  \begin{minipage}[c][3.5cm][c]{0.22\textwidth}
    \centering
    \includegraphics[width=0.9\linewidth]{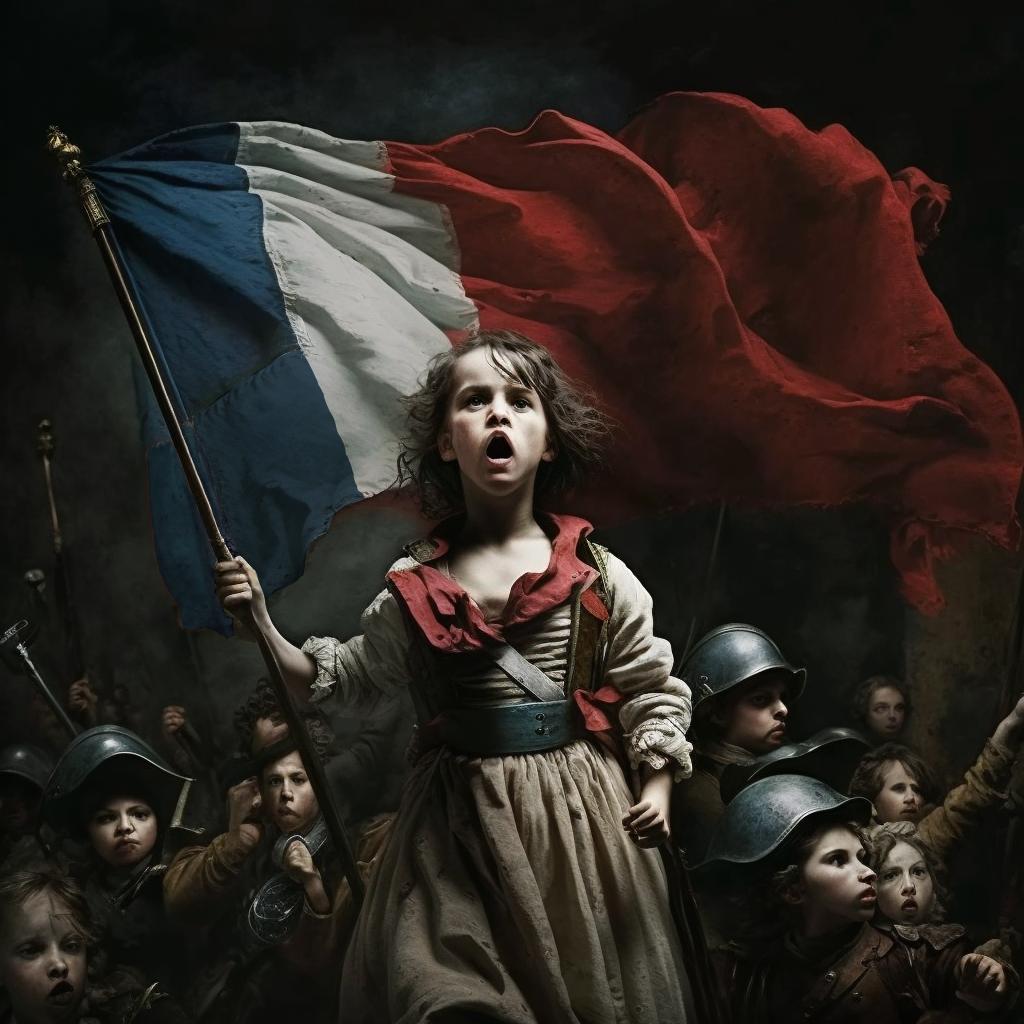}
  \end{minipage}
}
    & $0$   & {\footnotesize The image is a dramatic and evocative artwork depicting a young girl standing in the center, holding a flag with the colors of the French flag—blue, white, and red.} \\
    & $3$  & {\footnotesize The image depicts a young girl standing in the center, holding a small, tattered flag with the design of the French flag. 
} \\
    & $10$ & {\footnotesize The image shows a Telephone P p p p p p p p p p p p p p p p p p Telephone P p p p p p p p p p p p p p p p p p Telephone P p p p
} \\
\bottomrule
\end{tabularx}
\caption{\textbf{Effect of intervention strength on caption generation quality.} Examples of captions generated by LLaVA-NeXT under different intervention strengths ($\lambda = 0, 3, 10$). As intervention magnitude increases, captions begin to diverge from the original output. At moderate levels ($|\lambda| = 3$), outputs remain coherent but show lexical and structural variations. At high levels ($|\lambda| = 10$), generations often degrade into repetitive or nonsensical sequences. }
\label{examples_ref}
\end{table*}

\section{Details on the Number of Heads Selected and Control Experiment}
\begin{figure}[h]
\centering
\includegraphics[width=\linewidth]{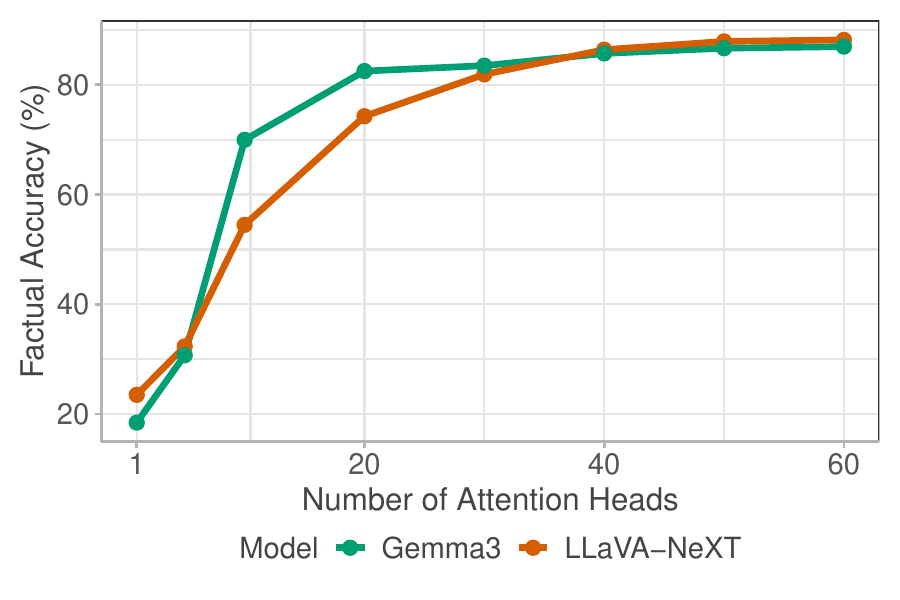}
\caption{\textbf{Effect of intervening on varying numbers of attention heads.} Change in factual accuracy as a function of the number of attention heads involved in the intervention. Each value $x$ indicates that $x$ heads are selected from both the factual and counterfactual groups. Intervention strength is fixed at $\lambda=3$. The results highlight that intervening on $20$ heads provides the optimal trade-off, maximizing factual accuracy without excessively affecting model stability.}
\label{fig:multik_heads_intervention}
\end{figure}
\begin{figure}[h]
\centering
\includegraphics[width=\linewidth]{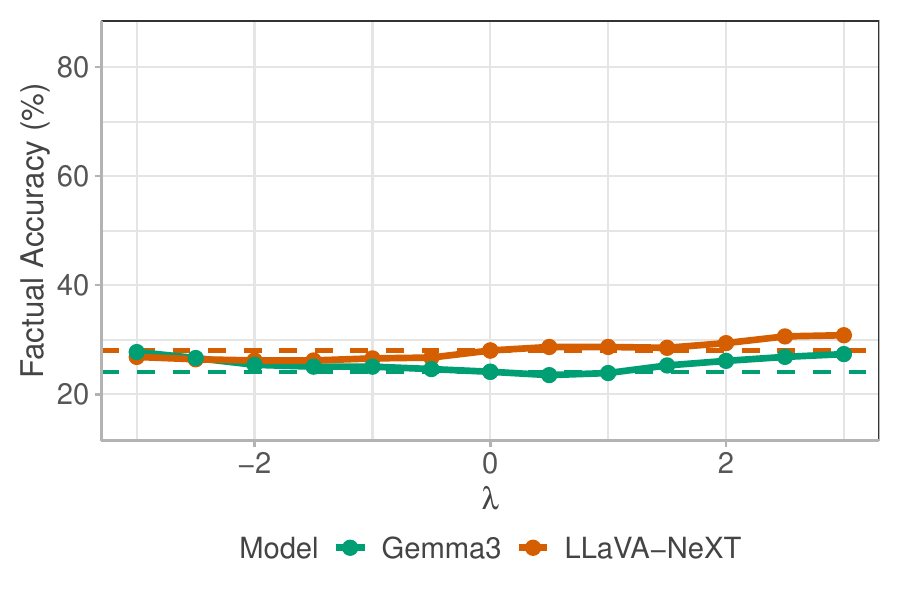}
\caption{\textbf{Control experiment: intervention on random attention heads.}  
Change in factual accuracy under varying levels of intervention strength ($\lambda$) applied to 100 randomly selected attention heads. The results show no substantial deviation from baseline, confirming the specificity of the identified target heads.}
\label{app:fig_additional}
\end{figure}
\Cref{fig:multik_heads_intervention} examines the effect of varying the number of intervened attention heads, with intervention strength fixed at $\lambda=3$. We observe that factual accuracy increases as the number of heads grows, reaching its peak at $20$ heads. Beyond this point, further interventions do not yield additional gains and may introduce instability. This demonstrates that intervening on $20$ heads provides the best balance between accuracy improvement and model robustness.
\label{app:n_heads}
\Cref{app:fig_additional} reports the control experiment, where the intervention was applied to $100$ randomly selected attention heads. The results show no measurable change in factual accuracy, confirming that improvements are not due to random head selection but to the specific heads identified in our method.
\label{app:additional}

\section{Details on the Intervention Parameter Choice}
\begin{figure}[h]
\centering
\includegraphics[width=\linewidth]{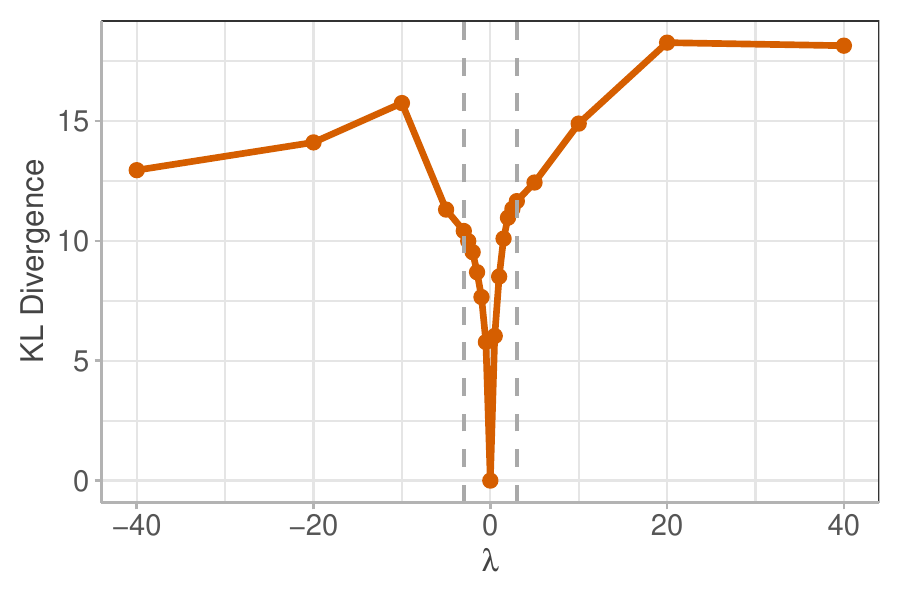}
\caption{\textbf{KL divergence between generated captions at different intervention strengths in LLaVA-NeXT.}
Symmetric increase in KL divergence around $\lambda=0$, with rapid divergence until $|\lambda|=3$ and stabilization near $|\lambda|=10$. Higher intervention magnitudes cause substantial shifts in the generated token distribution, indicating degradation in caption quality.}
\label{fig:kl_llava}
\end{figure}
\label{app:lambda_selection}
To ensure plausible interventions, we constrain the scaling parameter to $\lambda \in [-3, 3]$ and monitor the position of the higher-logit token in the full next-token distribution. 
For example, using LLaVA-NeXT, the average rank of the token $\tfact$ shifts from 3 at $\lambda = 0$ to 31 at $\lambda = 3$, indicating that while the intervention is highly effective, it introduces some deviation in the overall logit distribution, an expected effect when strongly modulating internal components.

To further support this choice of intervention range, we also prompt the model to generate captions with and without intervention, and manually inspect the quality of the outputs as we increase the intervention strength, $|\lambda|$. 
We empirically observe that for $|\lambda|$ greater than three, the quality of the generated captions degrades, and most of the time, they become agrammatical when $|\lambda|>10$ (see \cref{examples_ref}).

We also attempted to quantify the quality of the generated text after the intervention with a KL-divergence with the generated text before the intervention ($|\lambda|>0$), which we consider as a reference for a well-structured sentence. 
\Cref{fig:kl_llava} shows the average KL-divergence across all examples in \texttt{\textsc{WHOOPS-AHA!}} as we increase  $|\lambda|$ in LLaVA-NeXT. 

The KL divergence sharply increases for $|\lambda| < 3$, and then the growth slows down and stabilizes around $|\lambda|=12$ for $\lambda < -20$ and 18 for $\lambda > 20$.
When the KL is smaller than 10, for $\lambda$ between -3 and 3, the output sentences have a similar quality to those generated before intervention.


\section{Generalization to Visual CounterFact}
\label{app:generalization_vcf}

To evaluate whether the mechanisms for resolving factual and counterfactual conflicts identified with \texttt{\textsc{WHOOPS-AHA!}} generalize to other data distributions, we extended our analysis to the \texttt{Visual CounterFact} dataset \citep{Golovanevsky2025pixelVSpriors}.

\paragraph{Experimental setup.}
We utilized the ``color'' split of \texttt{Visual CounterFact}, which consists of paired factual and counterfactual images depicting the same object with different color attributes (e.g., a standard red strawberry versus a counterfactual blue strawberry). This dataset offers a distinct visual domain characterized by more photo-realistic manipulation compared to the illustrative nature of some \texttt{\textsc{WHOOPS-AHA!}} samples.

\begin{figure}[h!]
\centering
\includegraphics[width=\linewidth]{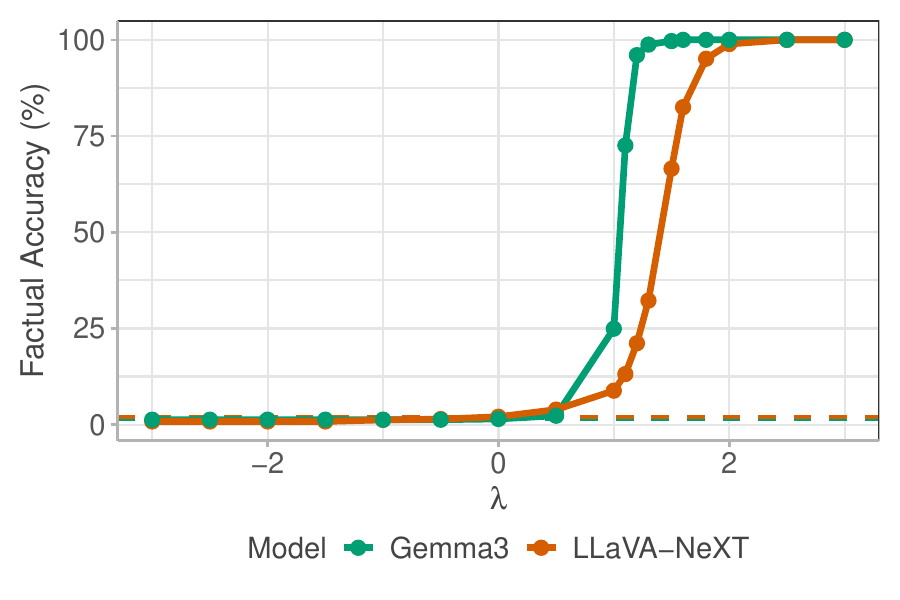}
\caption{\textbf{Effect of intervention on factual accuracy in \texttt{Visual CounterFact}.}
Positive values of $\lambda$ effectively steer both models toward factual accuracy, confirming that the results generalize to this dataset.}
\label{fig:pixelvsprior_intervention}
\end{figure}

\paragraph{Prompt construction.}
For each image, we constructed a completion prompt of the form: \texttt{"The color of the [object] is"}. We then analyzed the model's logits for the specific factual and counterfactual color tokens provided by the dataset (e.g., \textit{``red''} vs. \textit{``blue''}).

\paragraph{Procedure.}
We replicated the procedure described in \cref{sec:attention_heads} and \cref{sec:intervention}:
\begin{enumerate}
    \item We identified the top-20 factual and counterfactual attention heads specific to this dataset by analyzing the Logit Attribution on the final token position.
    \item We performed the attention intervention using the parameter $\lambda$ to steer the model between parametric knowledge (factual) and visual evidence (counterfactual).
\end{enumerate}

\paragraph{Intervention efficacy.}
\Cref{fig:pixelvsprior_intervention} illustrates the effect of intervention on LLaVA-NeXT and Gemma3. Consistent with our main results, increasing the activation of factual heads ($\lambda>0$) significantly boosts the model's reliance on internal parametric knowledge. Due to the naturally high success rate of the visual override in this dataset (low factual baseline), the effect of negative $\lambda$ is negligible, as the model is already effectively attending to the visual evidence.

\paragraph{Mechanism transferability.}
A key question is whether the specific attention heads mediating this conflict are consistent across datasets. We compared the top-20 heads identified on Visual CounterFact with those identified on \texttt{\textsc{WHOOPS-AHA!}}. We observed a strong overlap:
\begin{itemize}
\item $13$ out of $20$ counterfactual heads in LLaVA-Next and $14$ out of $20$ counterfactual heads in Gemma3 are identical.
\item $10$ out of $20$ factual heads are identical across the two models.
\end{itemize}
This substantial overlap suggests that the identified heads are not specific to \texttt{WHOOPS-AHA!}. Instead, they appear to reflect a stable and transferable mechanism for resolving conflicts between visual input and parametric knowledge across independently constructed datasets and varying degrees of photo-realism.

\section{Specificity of Conflict-Resolution Heads: Control on General Visual Understanding}
\label{app:pope}

A potential alternative explanation for the role of the identified counterfactual heads is that they function as generic vision-centric heads---that is, heads broadly responsible for processing visual information regardless of whether a knowledge conflict is present. Under this hypothesis, suppressing their image attention should degrade performance on any visually grounded task, not only on conflict-resolution ones.

To test this directly, we evaluate model performance on \texttt{\textsc{POPE}}~\citep{li2023pope}, a standard binary VQA benchmark consisting of 1{,}000 yes/no questions about object presence in images. \texttt{\textsc{POPE}} is a heavily vision-dependent task: removing the image entirely reduces accuracy to chance level (0.50) for both models, confirming that correct answers cannot be recovered from text alone.

We apply the same intervention used in our main analysis: we zero out the image attention of the identified counterfactual heads at the final token position (equivalent to $\lambda = 1$ applied to image tokens of $\mathcal{H}_\text{cofa}$), and measure accuracy before and after intervention. Results are reported in Table~\ref{tab:pope}.

\begin{table}[h]
\centering
\caption{\textbf{Effect of suppressing counterfactual head image attention on \texttt{\textsc{POPE}} accuracy.} Removing the image entirely collapses accuracy to chance (0.50), confirming the task is vision-dependent. Suppressing image attention only in the identified counterfactual heads leaves accuracy unchanged, indicating these heads are not involved in general visual processing.}
\label{tab:pope}
\resizebox{\columnwidth}{!}{%
\begin{tabular}{lccc}
\toprule
\textbf{Model} & \textbf{No image} & \textbf{Baseline} & \textbf{After intervention} \\
\midrule
Gemma3-12B    & 0.50 & 0.84 & 0.84 \\
LLaVA-NeXT-7B & 0.50 & 0.87 & 0.87 \\
\bottomrule
\end{tabular}
}
\end{table}

The results show no measurable degradation in \texttt{\textsc{POPE}} accuracy after suppressing image attention in the identified heads. This stands in sharp contrast to the no-image condition, where accuracy collapses to chance. The absence of any effect indicates that the identified heads are not required for general visual understanding, and therefore cannot be characterized as generic vision-centric components.

This finding reinforces our interpretation: the counterfactual heads do not attend to image tokens simply because they process visual input, but specifically because a competition between parametric knowledge and visual evidence is active. Their selective involvement under conflict conditions, combined with their irrelevance to routine visual processing, supports the claim that they constitute a dedicated mechanism for cross-modal conflict resolution.

\section{Quantitative Analysis of Visual Attribution}
\label{app:segmentation}

To validate the qualitative observation that counterfactual attention heads act as precise visual pointers, we performed a quantitative evaluation on a subset of $20$ sampled images. For each image, we obtained ground-truth object masks corresponding to the counterfactual concept (e.g., the "oranges" in the surgical scene) using the SAM3 model~\citep{sam3}.

\paragraph{Metric definition.}
We evaluate the alignment between the attention maps and the ground-truth masks using the \textit{Average Ratio}. This metric compares the attention intensity on the object versus the background. For a set of attention heads (e.g., the identified counterfactual heads), we first compute the average attention map across the heads for the given image. The ratio is then defined as:
\begin{equation}
\text{AvgRatio} = \frac{\frac{1}{|O|} \sum_{i \in O} \bar{A}i}{\frac{1}{|B|} \sum_{j \in B} \bar{A}_j}
\end{equation}
where $O$ is the set of visual tokens falling within the counterfactual object mask, $B$ is the set of background visual tokens, and $\bar{A}_i$  represents the attention value at visual token $i$ averaged across the selected $N=20$ heads. A value $>1$ indicates that the mechanism collectively focuses more intensely on the object than on the background.

\paragraph{Results.}
We compared our $20$ identified counterfactual attention heads against two baselines: (1) a gradient baseline, using standard gradient-based attribution of visual tokens that are responsible for the counterfactual prediction, and (2) a random baseline, obtained by 20 randomly sampled attention heads.

\begin{table}[h]
\centering
\caption{\textbf{Quantitative comparison of visual attribution methods on $20$ sampled images.} We report the median and IQR. The $p$-value indicates significance against the counterfactual heads method selection (Wilcoxon signed-rank test).}
\label{tab:quant_results}
\resizebox{\columnwidth}{!}{%
\begin{tabular}{llcc}
\toprule
\textbf{Model} & \textbf{Method} & \textbf{Average Ratio} & \textbf{$p$-value} \\
& & \small{(Median [IQR])} & \small{(vs. Ours)} \\
\midrule
\multirow{3}{*}{\textbf{Gemma 3}} 
& \textbf{Counterfactual heads} & \textbf{4.41} \small{[1.09 -- 7.85]} & -- \\
& Gradient & 1.74 \small{[1.09 -- 2.01]} & $<0.01$ \\
& Random heads & 0.92 \small{[0.71 -- 1.22]} & $<0.01$ \\
\midrule
\multirow{3}{*}{\textbf{LLaVA-NeXT}} 
& \textbf{Counterfactual heads} & \textbf{2.05} \small{[1.48 -- 3.33]} & -- \\
& Gradient & 1.88 \small{[1.26 -- 2.44]} & $0.003$ \\
& Random heads & 1.09 \small{[0.86 -- 1.25]} & $<0.01$ \\
\bottomrule
\end{tabular}%
}
\end{table}
\Cref{tab:quant_results} summarizes the results for both Gemma3 and LLaVA-NeXT. We report the Median and Inter-Quantile Range (IQR) as the distributions are non-normal. We also perform a Wilcoxon signed-rank test to assess whether the counterfactual heads assign significantly higher attention to the segmented regions than the baselines. In both architectures, the counterfactual heads significantly outperform the baselines, achieving a much higher contrast ratio.

For Gemma3, the identified counterfactual heads are highly selective with a median contrast of $4.41$, indicating that object patches receive over four times the attention intensity of background patches. 
For LLaVA-NeXT, the median ratio is $2.05$, meaning the object is attended to with double the intensity of the background.
In both cases, the counterfactual heads show a statistically significant improvement over the gradient and random baselines, confirming that the identified mechanism acts as a precise visual pointer across different VLM architectures.
\section{Prompts For Dataset Generation}

\label{app:prompt}

\begin{tcolorbox}[mycustombox={Prompt Used to Generate Dataset Instances.}]

You are a helpful assistant expert in LLMs research. 

Counterfactual Dataset Generation Prompt

Objective:
Generate captions for images that highlight a clear contrast between common (factual) and unusual (counterfactual) scenarios involving the subject depicted. Each caption must include the subject of the image and end with "\_\_\_" indicating the blank space where a single-word token is placed.

Definitions:
- **Factual token**: A single word that represents typical, expected behavior or attributes of the main subject shown in the image.
- **Counterfactual token**: A single word introducing a surprising, unexpected, or unusual element related explicitly to the same main subject; it makes sense only if the image explicitly illustrates this twist.

Context Provided:
For each image, you will receive the following textual information:
- Selected Caption: A primary description identifying the main subject clearly.
- Crowd Captions: Alternative descriptions from multiple annotators.
- Designer Explanation: Explanation emphasizing the unusual or counterintuitive aspect involving the subject.
- Crowd Explanations: Multiple explanations focusing on the unusual aspects related directly to the subject of the image.

Task Instructions:

Caption Construction:
- Create exactly one neutral sentence (caption) clearly containing the main subject depicted in the image, but avoiding the description of unusual aspects contained in the image.
- The sentence must end with an intentional blank ("\_\_\_").
- Critical Requirement: The caption must compel the model to complete the blank differently based on the context:
    - **Without the image**: complete with a factual token (typical scenario involving the subject).
    - **With the image**: complete with a counterfactual token (unexpected scenario explicitly depicted).
- Important Constraint: Use neutral language with NO textual hints indicating abnormality. The main subject must explicitly appear in the caption to establish context clearly. Only the image content itself should disambiguate the scenario.
- The caption should not contain any unusual or counterintuitive elements; the unusual aspect should be reflected solely in the image content and in the counterfactual tokens.
- Make sure that if you substitute the blank with a factual or counterfactual token, the sentence is fluent and grammatically correct.

Explicit Single-Word Token Generation:
- Generate exactly **ten single-word factual tokens** representing common scenarios involving the main subject that could complete in a grammatically correct way the sentence.
- Generate exactly **ten single-word counterfactual tokens** representing surprising scenarios involving the same subject, justified solely by the provided image, and that could complete the sentence in a grammatically correct way.
- Strictly enforce single-word tokens; no multi-word phrases or sentences.
- Ensure clear differentiation without conceptual overlap between factual and counterfactual tokens.

JSON Output Format:
Provide each caption and tokens following this exact schema:

\{
  "caption": "Neutral sentence explicitly containing the main subject and ending with an intentional blank ('\_\_\_')",
  "factual\_tokens": ["token1", "token2", "token3", "token4", "token5", ...],
  "counterfactual\_tokens": ["token1", "token2", "token3", "token4", "token5", ...],
  "context": \{
    "selected\_caption": "Primary description clearly stating the main subject of the image",
    "crowd\_captions": ["Caption 1", "Caption 2", "..."],
    "designer\_explanation": "Explanation highlighting the unusual aspect directly involving the main subject",
    "crowd\_explanations": ["Explanation 1", "Explanation 2", "..."]
  \}
\}

Your role is to craft neutral captions explicitly containing the main subject of each image, along with precisely differentiated factual and counterfactual single-word tokens. The explicit presence of the main subject in the caption must guide factual versus counterfactual completions, relying solely on the provided image for disambiguation.
\end{tcolorbox}

\begin{tcolorbox}[mycustombox={Prompt Used to Generate Factual and Counterfactual Tokens.}]

    You are presented with an image and an incomplete sentence describing its content. The image intentionally portrays an unusual scenario that contrasts typical or factual knowledge.

    Your task is to generate two lists of tokens:

    1. Factual Tokens (5 tokens): These tokens should represent words or concepts that accurately and typically complete the sentence based solely on common knowledge, without considering the unusual image.

    2. Counterfactual Tokens (5 tokens): These tokens should represent words or concepts that correctly complete the sentence when explicitly considering the unusual content depicted in the image, even if it contradicts common factual knowledge.

Please format your response clearly as a JSON object as follows:

```json
\{
  "sentence": "{INCOMPLETE\_SENTENCE}",
  "factual\_tokens": ["token1", "token2", "token3", "token4", "token5"],
  "counterfactual\_tokens": ["token1", "token2", "token3", "token4", "token5"]
\}
```

Choose tokens that clearly differentiate between typical knowledge and the unusual scenario depicted by the provided image.
\end{tcolorbox}
\end{document}